\documentclass[10pt,twocolumn,letterpaper]{article}

\usepackage{cvpr}              

\usepackage{cuted}
\usepackage{mdframed}
\usepackage{multirow}  %
\usepackage{makecell}

\usepackage{amsthm}

\PassOptionsToPackage{round}{natbib}
\usepackage[utf8]{inputenc} 
\usepackage[T1]{fontenc}    
\usepackage{url}            
\usepackage{booktabs}       
\usepackage{amsfonts}       
\usepackage{nicefrac}       
\usepackage{microtype}      
\usepackage{xcolor}         
\usepackage{arydshln}
\usepackage{colortbl}
\usepackage{algpseudocode}
\usepackage{algorithm}
\usepackage{amssymb} 
\usepackage{bm}
\usepackage{caption}
\usepackage{amsmath} 
\usepackage{graphicx}
\usepackage{subcaption}
\usepackage{bbm}
\usepackage{multirow}
\usepackage{enumitem}
\usepackage{minitoc}

\definecolor{mydarkred}{rgb}{0.6,0,0}
\definecolor{mydarkgreen}{rgb}{0,0.6,0}

\newtheorem{theorem}{Theorem}
\newtheorem{lemma}{Lemma}

\newtheorem{assumption}{Assumption}
\theoremstyle{remark}
\newtheorem{remark}{Remark}
\theoremstyle{definition}

\newcommand{\ignore}[1]{}
\definecolor{cvprblue}{rgb}{0.21,0.49,0.74}
\usepackage[pagebackref,breaklinks,colorlinks,allcolors=cvprblue]{hyperref}

\def\paperID{13345} 
\def\confName{CVPR}
\def\confYear{2026}

\title{Semi-Supervised Conformal Prediction With Unlabeled Nonconformity Score}

\author{Xuanning Zhou$^{1,2,\ddag,}$\thanks{Equal contribution. \  $\ddag$ Work done while working at SUSTech as a visiting scholar. \ $^\dag$ Corresponding author \{\texttt{weihx@sustech.edu.cn}\}.}
\quad
Zihao Shi$^{1,3,\ddag,*}$
\quad
Hao Zeng$^{1}$
\quad
Xiaobo Xia${^4}$
\quad
Bingyi Jing$^{2,5}$
\quad
Hongxin Wei$^{1,\dag}$\\
{\small $^1$Southern University of Science and Technology \hfill
$^2$The Chinese University of Hong Kong, Shenzhen \hfill
$^3$Fudan University \hfill} \\
{\small $^4$University of Science and Technology of China \qquad \quad 
$^5$Shenzhen Loop Area Institute}\\
}

\begin{document}
\maketitle


\begin{abstract}
    Conformal prediction (CP) is a powerful framework for uncertainty quantification, generating prediction sets with coverage guarantees.
    Split conformal prediction relies on labeled data  
    in the calibration procedure. 
    However, the labeled data is often limited in real-world scenarios, leading to unstable coverage performance in different runs. 
    To address this issue, we extend CP to the semi-supervised setting and propose SemiCP, a new paradigm that leverages both labeled and unlabeled data for calibration. 
    To achieve this, we introduce an unlabeled nonconformity score, Nearest Neighbor Matching (NNM) score. 
    Specifically, NNM estimates the nonconformity scores of unlabeled samples using their most similar pseudo-labeled counterparts during calibration, while maintaining the original scores for labeled data. 
    Theoretically, we demonstrate that the average coverage gap (i.e., the absolute difference between the empirical marginal coverage and the target coverage) 
    of SemiCP can decrease significantly at a rate $\mathcal{O}\bigl(1/\sqrt{N}\bigr)$ and converge to an error term, where $N$ is the number of unlabeled data. 
    Extensive experiments validate the effectiveness of SemiCP under limited labeled data, reducing the average coverage gap by up to 77\% on common benchmarks with 4000 unlabeled examples, when there are only 20 labeled examples. 
\end{abstract}
\begin{figure*}[t]
    \centering
    \begin{subfigure}[t]{0.54\textwidth}
        \centering
        \includegraphics[width=\linewidth]{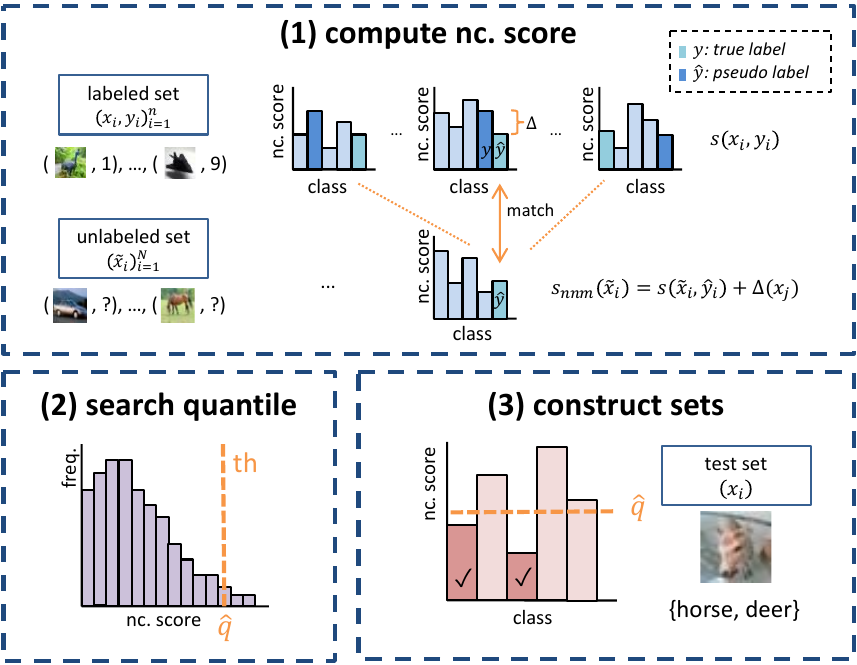}
        \caption{Method Overview}
        \label{fig:method}
    \end{subfigure}
    \begin{subfigure}[t]{0.33\textwidth}
        \centering
        \includegraphics[width=\linewidth]{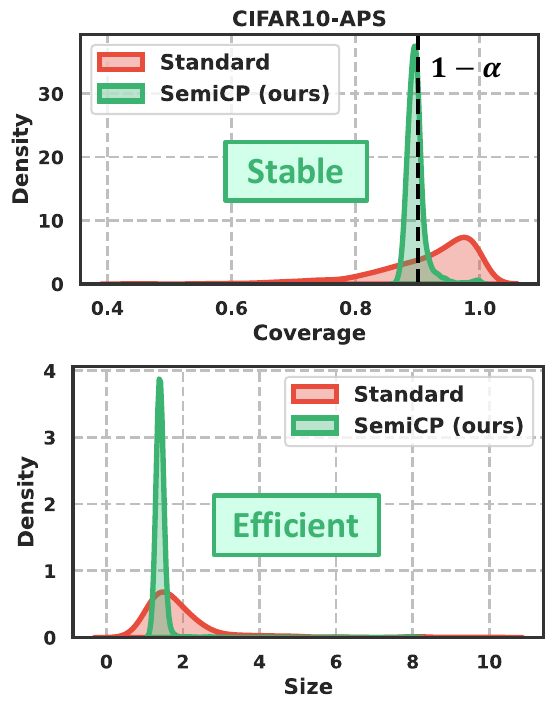}
        \caption{Experimental Results}
        \label{fig:results}
    \end{subfigure}
    \caption{Overview of SemiCP and its results. 
    (a) Method: use unlabeled scores via NNM scores to augment calibration, estimate the $(1-\alpha)$ quantile, and construct sets. 
    (b) Results: SemiCP concentrates coverage near the nominal $1-\alpha$ (closer is better) and yields smaller prediction sets (lower is better), mitigating the instability and over-sized sets of standard split CP.}
    
    \label{fig:method_results}
    \vspace{-3mm}
\end{figure*}

\section{Introduction}\label{sec:intro}
Uncertainty qualification is crucial for the safe deployment of machine learning models, particularly in high-stakes applications such as financial decision-making \citep{cresswell2024conformal, vovk2018conformal} and medical diagnostics \citep{vazquez2022conformal, olsson2022estimating}. 
This necessity highlights the utility of Conformal Prediction (CP), a statistical framework that produces prediction sets containing ground-truth labels with coverage guarantee \citep{vovk2005algorithmic, shafer2008tutorial, angelopoulos2023conformal}. 
In particular, smaller \textit{valid} prediction sets indicate lower uncertainty in the predictions, and vice versa. 
In this manner, conformal prediction can communicate the degree of trustworthiness of model outputs in a transparent and interpretable way.

Split Conformal Prediction \citep{papadopoulos2002inductive, lei2018distribution}, a popular CP paradigm, utilizes a labeled hold-out set to calibrate the threshold for constructing the prediction set. 
While providing valid prediction sets, this framework usually suffers from 
instability in different runs when labeled data are limited (see Fig.~\ref{fig:pred_cov_kde}) \citep{linusson2014efficiency, ding2023class}.
To alleviate this issue, previous works attempted to interpolate calibration instances or employ a modified p-value, but they are heuristic and do not provide a finite-sample guarantee \citep{johansson2015handling, carlsson2015modifications}. 
Recently, few-shot CP \citep{fisch2021few} employs a meta-learning approach using similar auxiliary tasks, but its dependence on exchangeable task collections limits practicality.
This motivates us to explore another alternative -- leveraging unlabeled data to address the data-scarce problem of labeled calibration sets. 

In this work, we propose Semi-Supervised Conformal Prediction (SemiCP), a paradigm that leverages both labeled and unlabeled data to calibrate conformal predictors  (see Fig.~\ref{fig:method_results}). 
To effectively utilize unlabeled examples, we introduce a new nonconformity score, \emph{Nearest Neighbor Matching} (NNM), which estimates unlabeled scores by aligning them with the most similar labeled instances in the pseudo score space. 
This design enables SemiCP to approximate the true score distribution, allowing us to leverage unlabeled data to stabilize the quantile estimation. 
Theoretically, we demonstrate that SemiCP shrinks the average coverage gap (i.e., the absolute difference between the empirical marginal coverage and the target coverage) at a rate $\mathcal{O}(1/\sqrt{N})$, where $N$ is the number of unlabeled samples. 
Notably, our SemiCP with NNM scores offers a general solution to the instability probelem under limited labeled data, and is complementary to existing conformal prediction methods.

To validate our method, we conducted extensive experiments on three image classification datasets: CIFAR-10, CIFAR-100~\citep{krizhevsky2009learning}, and ImageNet~\citep{deng2009imagenet}.
First, we empirically show that our method consistently enhances the stability and efficiency of existing scoring functions, including THR \citep{sadinle2019least}, APS \citep{romano2020classification}, and RAPS \citep{angelopoulos2020uncertainty}. 
For example, on CIFAR-10 with 20 labeled data, SemiCP narrowed the average coverage gap by 77\% and the prediction set size by 5.7\% with 4000 unlabeled data. 
Moreover, our SemiCP can be applied effectively to conditional CP~\citep{papadopoulos2002inductive, vovk2005algorithmic} and integrated with other CP algorithms (such as ClusterCP \citep{ding2023class}). 
Finally, our SemiCP works well with different model architectures, highlighting its robustness across diverse backbones. 
In summary, these results collectively demonstrate the remarkable effectiveness, scalability, and broad applicability of the SemiCP for reliable uncertainty quantification. 




We summarize our contributions as follows:
\begin{itemize}[leftmargin=7pt]
    \item We introduce SemiCP, a semi-supervised conformal prediction paradigm, and the \emph{Nearest Neighbor Matching} (NNM) scores to enhance the stability in standard Conformal Prediction under limited labeled data.
    \item We provide a theoretical analysis by proving that adding unlabeled samples in SemiCP results in the decrease of the average coverage gap. This theoretical insight shows that including unlabeled data
    enhances the stability of CP.
    \item We conduct extensive experiments that demonstrate effectiveness and applicability across datasets and models. 
    The method is data-efficient and robustness, with ablations and sensitivity analysis detailed in the Appendix~\ref{sec:analysis} \& \ref{sec:exp}. 
\end{itemize}


\section{Related work}\label{sec:related_work}
\paragraph{Conformal prediction.}
Conformal Prediction \citep{vovk2005algorithmic, lei2018distribution, papadopoulos2002inductive} provides a model-agnostic, finite-sample, and distribution-free framework for constructing prediction sets with coverage guarantee, and has been applied across various domains such as classification, regression \citep{romano2019conformalized, xi2025robust}, and large language models \citep{su2024api, cherian2024large}. 
In this work, we focus on the Split Conformal Prediction framework \citep{angelopoulos2023conformal}, where the training and calibration sets are separate. 
A key direction of research is conditional coverage, which involves obtaining coverage guarantee under specific conditions, such as when $X=x$ (\textit{X}-conditional, usually relaxed to group-conditional) \citep{vovk2005algorithmic, gibbs2025conformal} or $Y=y$ (\textit{Y}-conditional, or class-conditional) \citep{shi2013applications, lofstrom2015bias}, allowing more contextually relevant prediction sets. Another important direction is the design of nonconformity scores, such as THR \citep{sadinle2019least}, APS \citep{romano2020classification}, RAPS \citep{angelopoulos2020uncertainty}, and SAPS \citep{huang2023conformal}, which aim to improve the efficiency of prediction sets. 
However, current uncertainty scores generally require a certain amount of ground-truth labeled data in the calibration set.
Our work is complementary to these labeled nonconformity scores and can be generalized to conditional coverage settings.

In conformal prediction, 
the dependence on a large calibration set has been a longstanding concern. 
\citet{linusson2014efficiency} analyze the effect of the calibration size and recommend at least several hundred examples, a requirement that is often infeasible—especially for conditional guarantee. 
For example, on ImageNet with 1,000 classes, using 100 calibration samples per class requires \(10^{5}\) labeled examples.
To mitigate small-sample issues, prior work interpolates among calibration instances \citep{johansson2015handling}, clusters points to share the calibration mass (Cluster-CP; \citealp{ding2023class}), or applies meta-learning in few-shot CP \citep{fisch2021few}. However, none of these approaches uses unlabeled data to improve calibration. In this work, we first show that incorporating unlabeled samples can reduce the reliance on a large calibration set, thereby boosting the performance of CP under limited labeled data.

A parallel line of work \citep{mazuelas2025split} proposes unsupervised calibration by estimating label weights for unlabeled calibration examples via IPM minimization (RKHS/Lipschitz), but it entails optimizing an \(N\times K\) weight matrix (number of unlabeled samples \(\times\) classes), which can be computationally prohibitive and may even be intractable in large-sample. 
In contrast, our method requires neither access to the training dataset nor any additional training or optimization (training-free). SemiCP can directly compute nonconformity scores on unlabeled samples, which is easier to implement and use.

\paragraph{The benefits of unlabeled data.}
Some works have leveraged Conformal Prediction (CP) as a tool to address challenges in other semi-supervised learning domains. For example, CPSSDS~\citep{tanha2022cpssds} applies CP to self-training on data streams to tackle label scarcity and concept drift, while CSForest~\citep{han2024conformalized} integrates CP into the construction of random forests for anomaly detection. 
Our work is more closely related to research on semi-supervised predictive inference, in which unlabeled data are used to improve the performance of inference \citep{angelopoulos2023prediction, angelopoulos2023ppi++, zrnic2024cross, cortes2008sample, chakrabortty2022semi, kim2025semi}. 
Specifically, prediction-powered inference (PPI) \citep{angelopoulos2023prediction} showed that leveraging accurate model predictions can further sharpen confidence intervals, a strategy subsequently applied in conformal risk control \citep{einbinder2025semi}, causal inference \citep{gao2024role}, distribution learning \citep{wen2025semi}, and beyond. 
To the best of our knowledge, there is currently no existing work on CP in the semi-supervised setting, leveraging labeled data to estimate the true nonconformity socres of unlabeled data. 
In this paper, we present the first method for leveraging unlabeled data to improve the stability and efficiency of conformal prediction. A more detailed comparison between SemiCP and Semi-supervised quantile estimation can be found in the Appendix~\ref{sec:compare}.

\section{Preliminaries}\label{sec:preli}
Conformal prediction (CP)~\citep{vovk2005algorithmic} aims to produce prediction sets that contain ground-truth labels with the desired coverage rate. Here, we focus on the classification task.
Let $\mathcal{P}_{\mathcal{X}\mathcal{Y}}$ denote the joint distribution over inputs and labels, where $\mathcal{X} \subset \mathbb{R}^d$ denotes the input space, and $\mathcal{Y} = \{1, \dots, K\}$ denotes the label space in multi-class classification.
Formally, the goal of conformal prediction is to construct a set-valued mapping $\mathcal{C}: \mathcal{X} \rightarrow 2^{\mathcal{Y}}$ that satisfies a coverage probability near a target value $1-\alpha$
for a user-defined miscoverage level $\alpha \in (0,1)$. 
For conditional conformal prediction, the goal is to satisfy $
\mathbb{P}(y_{\text{test}} \in \mathcal{C}_{1-\alpha}(\boldsymbol{x})|\boldsymbol{x}\in G_i) \geq 1 - \alpha, \forall G_i \in \mathcal{G}$, where \(G_i\) denotes predefined subgroups.


As a popular CP framework, split conformal prediction~\citep{shafer2008tutorial} introduces a hold-out calibration set $\mathcal{D}_{\rm cal} = \{(\boldsymbol{x}_i, y_i)\}_{i=1}^n$ consisting of labeled examples drawn from $\mathcal{P}_{\mathcal{X}\mathcal{Y}}$, which are exchangeable. 
For each example $(\boldsymbol{x}_i, y_i)$ from the calibration set, we calculate the nonconformity score $s_i=S(\boldsymbol{x}_i,y_i)$ with a score function $S: \mathcal{X}\times\mathcal{Y}\rightarrow\mathbb{R}$, which quantifies how conformal a data point is. For \(q\in[0,1]\) and a finite set \(A\subseteq \mathbb{R}\), 
let \(\text{Quantile}(A, q)\) denote the smallest \(a\in A\) such that the fraction \(q\) of elements in \(A\) is less than or equal to \(a\). Then, we obtain a threshold as follows:
\begin{equation}\label{eq:tau}
    \hat{\tau} = \text{Quantile}\big(\{s_i\}_{i=1}^n, \frac{\lceil (n+1)(1-\alpha)\rceil}{n} \big).
\end{equation}

Given a test instance \(\boldsymbol{x}_\text{test}\), we compute the nonconformity score for each label $y \in \mathcal{Y}$ and construct the prediction set with the threshold $\hat{\tau}$ as
\[
   \mathcal{C}_{1-\alpha}(\boldsymbol{x}_\text{test}
   ) :=\{y\in\mathcal{Y}: S(\boldsymbol{x}_\text{test},y)\leq \hat{\tau} \}.
\]
The obtained prediction set $\mathcal{C}_{1-\alpha}(\boldsymbol{x}_\mathrm{test}
)$ satisfies the following finite-sample coverage guarantee:
$$1 - \alpha\le\mathbb{P}(y \in \mathcal{C}_{1-\alpha}(\boldsymbol{x}_\mathrm{test}
))\le1-\alpha +\frac{1}{n+1}$$
with the assumption of exchangeability \citep{lei2018distribution}. Of course, we can replace the assumption of exchangeability with sampling i.i.d. from the same distribution, which is also the assumption we will use in the following analysis.


\begin{figure}[t]
    \centering
    \includegraphics[width=\linewidth]{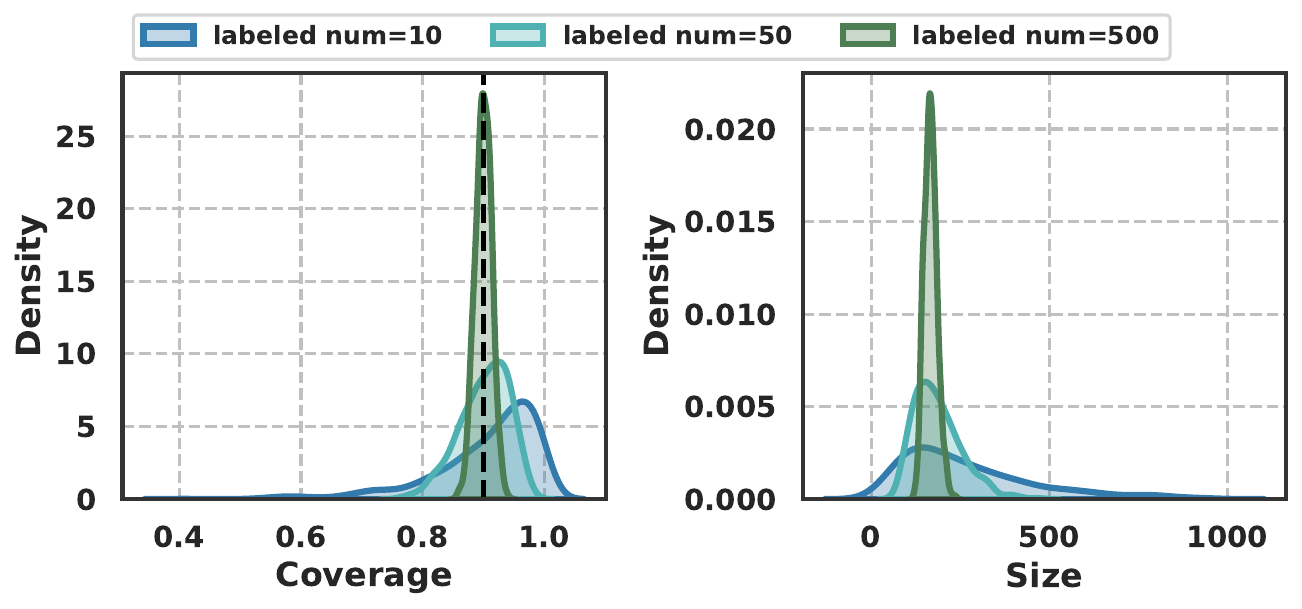}
    \vspace{-5mm}
    \caption{Kernel density estimates of coverage and prediction–set size as the number of labeled calibration points varies. Results are on ImageNet using APS, averaged over 100 independent runs. As the labeled calibration size grows, coverage concentrates around the nominal $1-\alpha$ and set sizes decrease, indicating improved stability and efficiency.}
    \label{fig:pred_cov_kde}
    \vspace{-3mm}
\end{figure}
\vspace{-3mm}
\paragraph{The issue of limited labeled data} 
Limited labeled data in calibration sets will lead to unstable coverage performance in different runs.
To illustrate this, we begin with a theoretical analysis. 
Assuming the nonconformity scores are continuous, the coverage given the calibration set follows a Beta distribution \citep{vovk2012conditional, angelopoulos2023prediction, ding2023class}:
\[\mathbb{P}(y_{\text{test}} \in \mathcal{C}_{1-\alpha}(\boldsymbol{x}_{\text{test}}) \mid \mathcal{D}_{\text{cal}}) \sim \mathrm{Beta}(l_\alpha^n, n+1 - l_\alpha^n),\]
where $l_\alpha^n = \lceil(n+1)(1-\alpha)\rceil$, $n$ is the number of labeled calibration data. 
The variance of the coverage given the calibration set is
\[\frac{l_\alpha^n (n+1 - l_\alpha^n)}{(n+1)^2 (n+2)} \approx \frac{\alpha(1-\alpha)}{n + 2},\]
which can be large when $n$ is small.
In particular, a small calibration set results in the coverage with high variability, leading to divergent results across different runs\citep{linusson2014efficiency, vovk2012conditional, foygel2021limits, lei2014distribution}.
For example, if $n=10$ and $\alpha=0.1$, then $\ell_{\alpha}^n=10$ and the real marginal coverage overshoots $1-\alpha$ by about 0.066, yet there remains a 10.7\% probability that coverage falls below 80\%.
The empirical results also provide strong evidence to support this conclusion.
We conducted experiments with varying numbers of labeled examples in calibration sets on ImageNet~\citep{deng2009imagenet} with $\alpha=0.1$. 
Fig.~\ref{fig:pred_cov_kde} shows the KDEs of coverage (left) and prediction‐set size (right) as the number of labeled calibration points decreases . As $n$ shrinks, the coverage distribution shifts upward (systematic over‐coverage) and widens markedly around the target $1-\alpha$ (vertical dashed line), while the size distribution becomes heavier‐tailed. The upward shift implies inflated prediction sets and reduced efficiency; increased dispersion indicates run‐to‐run instability, with a substantial chance of missing nominal coverage. 
We further quantify this instability as the \textit{Coverage Gap}, defined as $\Delta = \frac{1}{m}\sum_{i=1}^{m} |c_i - (1-\alpha)|$, where $c_i$ represents the empirical marginal coverage of the $i$-th run and $m$ is the total number of runs. 
A larger coverage gap corresponds to a larger deviation of marginal coverage in multiple runs, which means the performance of CP is more unstable. 
In general, limited calibration data make the performance of CP unstable, with larger sets and higher coverage gap that reduce reliability and interpretability.

Although split conformal prediction is highly effective in practice, the instability induced by small calibration sets is non-negligible. 
This motivates us to leverage unlabeled data to enlarge the calibration sets, thereby improving the stability of the resulting sets. 
In the next section, we introduce our semi-supervised paradigm, SemiCP, which incorporates unlabeled examples to stabilize the performance.

\section{Method}

In this section, we propose SemiCP, a semi-supervised conformal prediction framework that leverages both labeled and unlabeled data to address the instability and inefficiency issues under limited calibration samples. In Section~\ref{sec:meta-algorithm}, we introduce SemiCP and formalize the conditions under which its coverage guarantee holds when incorporating unlabeled scores. Then, in Section~\ref{sec:UNS}, we present \emph{NNM} -- an unlabeled nonconformity score function that can lessen instability and inefficiency under limited sampled data conditions, with valid coverage proved theoretically. We provide overall algorithm in Appendix~\ref{sec:sup_al}.

\subsection{Semi-supervised conformal prediction}\label{sec:meta-algorithm}
In the semi-supervised setting, we consider a calibration dataset $\mathcal{D}=\mathcal{D}_{\text{labeled}}\cup \mathcal{D}_{\text{unlabeled}}$ comprising both labeled and unlabeled examples: $\mathcal{D}_{\text{labeled}}=\{(\boldsymbol{x}_i, y_i)\}_{i=1}^n$ and $\mathcal{D}_{\text{unlabeled}}=\{(\tilde{\boldsymbol{x}}_i)\}_{i=1}^N$, which are i.i.d. sampled from $\mathcal{P}_{\mathcal{X}\mathcal{Y}}$. 
We denote by $\tilde{y}_i$ the unknown true label of the unlabeled instance $\tilde{\boldsymbol{x}}_i$. 
In SemiCP, we compute nonconformity scores for both labeled and unlabeled examples in the calibration set. For each labeled example $(\boldsymbol{x}_i, y_i) \in \mathcal{D}_{\text{labeled}}$, we use commonly used nonconformity score functions $S(\cdot)$, e.g. THR \citep{sadinle2019least}, APS \citep{romano2020classification} or RAPS \citep{angelopoulos2020uncertainty}. 
For each unlabeled instance, we compute its nonconformity score $\tilde{s}_i = \tilde{S}(\tilde{x}_i)$ with a specially-designed unlabeled score function $\tilde{S}(\cdot)$ that does not require ground-truth labels. We will introduce our design of the unlabeled score function in Section~\ref{sec:UNS}. 


Then, similar to Eq.~(\ref{eq:tau}) in split conformal prediction, we aggregate the nonconformity scores of both labeled and unlabeled examples to estimate the threshold $\hat{\tau}_\text{SemiCP}$: 
{
\small
\begin{equation}\label{eq:semicp}
\hat{\tau}_\text{SemiCP} = \text{Quantile}\left(\{\tilde{s}_i\}_{i=1}^N\cup\{s_i\}_{i=1}^n, \frac{\lceil (n+N+1)(1-\alpha)\rceil}{n+N} \right). 
\end{equation}
}

Finally, given the estimated threshold $\hat{\tau}_\text{SemiCP}$, the prediction set for a test input $\boldsymbol{x}_\text{test}$ can be constructed as:
$\mathcal{C}_\text{SemiCP}(\boldsymbol{x}_\text{test}; \hat{\tau}_\text{SemiCP}) = \{y\in\mathcal{Y}: S(\boldsymbol{x}_\text{test},y)\leq \hat{\tau}_\text{SemiCP} \}.$ 
With a carefully designed unlabeled score function $\tilde{S}(\cdot)$, SemiCP increases the nonconformity score calibration pool  with unlabeled examples, thus stabilizing threshold estimation and producing smaller and more efficient prediction sets. We now formalize the coverage properties of SemiCP.

\begin{theorem}[SemiCP]\label{thm:coverage}
    Assuming labeled data \(\{(\boldsymbol{x}_i, y_i)\}_{i=1}^n\) and unlabeled data \(\{\tilde {\boldsymbol{x}}_i\}_{i=1}^N\) be i.i.d. from $\mathcal{P}_{\mathcal{X}\mathcal{Y}}$. 
    Let \(F_{\tilde{S}}\) and \(F_S\) denote the cumulative distribution functions of the estimated and true nonconformity scores as
    \(
    F_{\tilde{S}}(t)=\mathbb{P}\bigl(\tilde S(\tilde{ \boldsymbol{x}})\le t\bigr), \
    F_{S}(t)=\mathbb{P}\bigl(S(\tilde{ \boldsymbol{x}}, \tilde y)\le t\bigr)
    \) for all \(t\in\mathbb{R}\).
    Then, for any target coverage $1-\alpha$, the SemiCP prediction set $\mathcal{C}_{\text{SemiCP}}$ satisfies the marginal coverage guarantee
    $$\mathbb{P}\bigl(y_{\mathrm{test}}\in\mathcal C_{\mathrm{SemiCP}}(\boldsymbol{x}_{\mathrm{test}})\bigr)\ge1-\alpha+\epsilon_{n,N},$$
    where $\epsilon_{n,N}=\frac{N}{N+n}(F_S(\hat{\tau}_\text{SemiCP})-F_{\tilde{S}}(\hat{\tau}_\text{SemiCP}))$.
\end{theorem}

We provide the proof in Appendix~\ref{sec:proof_thm1}. By Theorem~\ref{thm:coverage}, the incorporation of unlabeled scores can introduce a coverage bias proportional to the discrepancy between \(F_S\) and \(F_{\tilde S}\). 

The following theorem proves that the average coverage gap of SemiCP shrinks while adding unlabeled examples. 

\begin{theorem}[Calibration-conditional SemiCP]\label{thm:distribution}
    Given the calibration set $\mathcal{D}$. 
    Assume $\{(\boldsymbol{x}_i,y_i)\}_{i=1}^n$ and $\{(\tilde{\boldsymbol{x}}_i,\tilde y_i)\}_{i=1}^N$ are i.i.d. from the same distribution.
    Then, for any $\alpha,\delta\in(0,1)$, with probability at least $1-\delta$,
    \begin{align*}
        1-\alpha &+ \epsilon_{n,N}-\sqrt{\frac{\text{log}(\frac{2}{\delta})}{2(n+N)}}\le\mathbb{P}(y_{\text{test}} \in \mathcal{C}_{\text{SemiCP}}(\boldsymbol{x}_{\text{test}})|\mathcal{D})\\
        &\le1-\alpha+\epsilon_{n,N}
        +\frac{1}{n+N+1}
        +\sqrt{\frac{\text{log}(\frac{2}{\delta})}{2(n+N)}},
    \end{align*}
    where 
    $\epsilon_{n,N}=\frac{N}{N+n}(F_S(\hat{\tau}_\text{SemiCP})-F_{\tilde{S}}(\hat{\tau}_\text{SemiCP}))$.
\end{theorem}

We provide the proof in Appendix~\ref{sec:proof_thm2}.
From Theorem~\ref{thm:distribution}, the coverage gap shrinks at rate $\mathcal{O}\!\bigl(1/\sqrt{N}\bigr)$, while the bias term approaches $F_S(\hat{\tau}_{\mathrm{SemiCP}})-F_{\tilde{S}}(\hat{\tau}_{\mathrm{SemiCP}})$.
Thus, adding unlabeled data improves stability at the potential cost of a small coverage bias.
The accuracy of coverage depends on both $N$ and the discrepancy between $\tilde{S}$ and $S$; with an appropriate unlabeled score $\tilde{S}$, this bias can be made negligible.
In what follows, we introduce a tailored unlabeled nonconformity score that mitigates instability and inefficiency under limited labeled data while preserving valid coverage. 

\subsection{Nonconformity score for unlabeled data}\label{sec:UNS}

Estimating nonconformity scores for unlabeled data is difficult because the scores depend on unknown true labels—accurate estimation is essentially equivalent to recovering all unlabeled labels, i.e., requiring an oracle. 
Instead, we aim to approximate the score distribution: construct \(\tilde S\) so that \(F_{\tilde{s}}\) closely matches \(F_s\). 
By Theorem~\ref{thm:distribution}, this alignment controls the bias term \(\epsilon_{n,N}\) while preserving nominal coverage, thereby yielding a small coverage gap with a valid guarantee.
In the following, we introduce our unlabeled nonconformity score \textbf{Nearest Neighbor Matching (NNM)} that converges to the true scores in distribution asymptotically.

Let $f$ be a pre-trained classifier and $f_j(\tilde{\boldsymbol{x}}_i)$ the softmax probability for class $j$ on an unlabeled instance $\tilde{\boldsymbol{x}}_i$. The pseudo-label is
\(
\hat y_i \;=\; \arg\max_{j\in\mathcal{Y}} f_j(\tilde{\boldsymbol{x}}_i).
\)
A natural way to obtain unlabeled nonconformity scores is to plug $\hat y_i$ into an existing score function $S(\cdot)$ (e.g., THR, APS). We call the resulting value the \emph{pseudo score} and the procedure the \emph{naive method}:
\(
\tilde{S}_{\mathrm{naive}}(\tilde{\boldsymbol{x}}_i; S, f) \;=\; S(\tilde{\boldsymbol{x}}_i,\hat y_i).
\)
However, this estimator is systematically biased because $\hat y_i$ is the model’s most confident class, yielding artificially low nonconformity scores. Consequently, the empirical quantile $\hat\tau$ is underestimated and coverage is reduced. We define the \emph{pseudo bias} as the gap between the true and pseudo scores:
\[
\Delta(\tilde{\boldsymbol{x}}_i)\;:=\; S(\tilde{\boldsymbol{x}}_i,\tilde y_i)\;-\;S(\tilde{\boldsymbol{x}}_i,\hat y_i).
\]
Since the pre-trained model encodes our best available information about the unknown labels, we estimate the pseudo bias for unlabeled data by leveraging a small labeled set. Labeled examples allow us to relate pseudo scores to their observed biases and use this relation to debias unlabeled pseudo scores. 
For each unlabeled $\tilde{\boldsymbol{x}}_i$, we match a labeled example $\boldsymbol{x}_j$ whose \emph{pseudo score} is closest to that of $\tilde{\boldsymbol{x}}_i$:
\[
j \;=\; \text{argmin}_{j\in\{1,\ldots,n\}}
\bigl|\, S(\tilde{\boldsymbol{x}}_i,\hat y_i)\;-\;S(\boldsymbol{x}_j,\hat y_j)\,\bigr|.
\]

This nearest-neighbor matching in pseudo-score space provides a data-driven estimate of the local pseudo bias to correct the unlabeled pseudo score. Then, with the approximated bias, we compute the NNM score of an unlabeled example $\tilde{\boldsymbol{x}}_i$ by debiasing its pseudo score $S(\tilde{\boldsymbol{x}}_i, \hat{y}_i)$:
\begin{equation}\label{eq:nnm}
\begin{aligned}
    \tilde{S}_\mathrm{nnm}(\tilde{\boldsymbol{x}}_i&, D_{\mathrm{labeled}}, S, f)
    = S(\tilde{\boldsymbol{x}}_i, \hat{y}_i) + \Delta(\boldsymbol{x}_j) \\
    &= S(\tilde{\boldsymbol{x}}_i, \hat{y}_i) + S(\boldsymbol{x}_j, y_j) - S(\boldsymbol{x}_j, \hat{y}_j),
\end{aligned}
\end{equation}
where $S(\cdot)$ can be any existing labeled score function. 

\begin{figure}[t]
    \vspace{-2mm}
    \centering
    \includegraphics[width=0.8\linewidth]{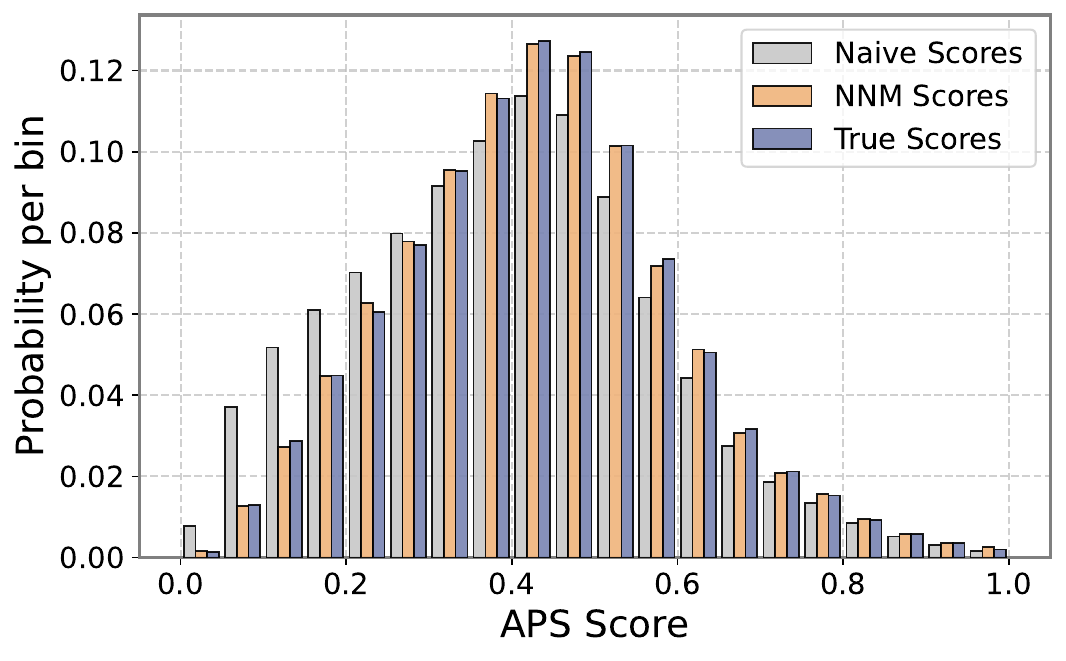}
    \vspace{-3mm}
    \caption{Empirical PDFs of nonconformity scores: naive vs.\ NNM vs.\ true. 
    Results on ImageNet with APS, aggregated over 100 runs. 
    The naive pseudo-label scores are systematically biased relative to the true scores, while NNM closely matches true distribution.}
    \label{fig:aps_dist}
    \vspace{-3mm}
\end{figure}

\paragraph{Empirical analysis.}
Empirically, we find that pseudo bias distributions of labeled and unlabeled examples with similar pseudo scores are nearly identical, supporting our nearest-neighbor assumption. 
Fig.~\ref{fig:aps_dist} shows that the empirical PDF of NNM scores closely matches the true PDF, whereas naive pseudo scores are consistently biased toward lower values. 
This distributional alignment leads to more accurate quantile estimation and substantially reduces average coverage gap when calibration data are scarce. 
Consequently, the SemiCP threshold $\hat{\tau}_{\mathrm{SemiCP}}$ obtained with NNM scores provides more stable coverage guarantees than standard split-CP while maintaining coverage valid.

\paragraph{Theoretical guarantee.}
To formalize the intuition that NNM approximates the true nonconformity distribution, we provide the following theoretical result. 
It characterizes how the cumulative distribution function (CDF) of the estimated unlabeled scores $F_{s_{\text{nnm}}}$ concentrates around that of the true scores $F_s$, under mild regularity assumptions. 
Intuitively, the parameters $\delta$, $\epsilon$, and $E$ control the deviation between the two CDFs, quantifying the bias introduced by approximate pseudo-labels and the variance reduction achieved through additional unlabeled calibration samples. 
As $\delta \!\to\! 0$ and the number of labeled examples $n$ increases, the bias term $(1-\epsilon)^n$ vanishes and $F_{s_{\text{nnm}}}$ converges to $F_s$, ensuring asymptotic consistency of the proposed NNM score. 

\begin{theorem}\label{thm:nnm_asy}
    Suppose $\delta>0$.
    Suppose Assumption \ref{asp1} 
    and Assumption \ref{asp2} 
    hold, then there exists $\epsilon, E\in(0,1)$ depending on $\delta$, and $A, B, M\in\mathbb{R}^+$, which satisfies for any $t$,
    \begin{align*}
        \bigl|F_{s_{\text{nnm}}}(t)-F_{s}(t)\bigr|\le G, 
    \end{align*}
    where $G=A\bigl[\bigl(1-(1-E)^n\bigr)2M\delta+B(1-\epsilon)^n\bigr]$.
\end{theorem}


The two assumptions required and the proof of the theorem are provided in Appendix~\ref{sec:proof_thm3}. 
As the bounds in Theorem~\ref{thm:nnm_asy} hold for all $t$ in the range of the nonconformity scores, the bounds are uniform in $t$ and can be directly plugged into Theorems~\ref{thm:coverage} and \ref{thm:distribution}. 
This theorem implies that NNM asymptotically preserves the true nonconformity score distribution as the labeled sample size grows, with the approximation bias bounded by $\delta$ and convergence rate controlled by $n$. 
Together with the empirical evidence in Fig.~\ref{fig:aps_dist}, this result establishes both theoretical and practical validity of our approach: NNM achieves lower coverage gap and tighter coverage with negligible degradation in coverage guarantees, making SemiCP a principled and robust semi-supervised extension of conformal prediction. 


\begin{figure*}[t]
    \centering
    \begin{subfigure}{0.3\textwidth}
        \centering
        \includegraphics[width=\textwidth]{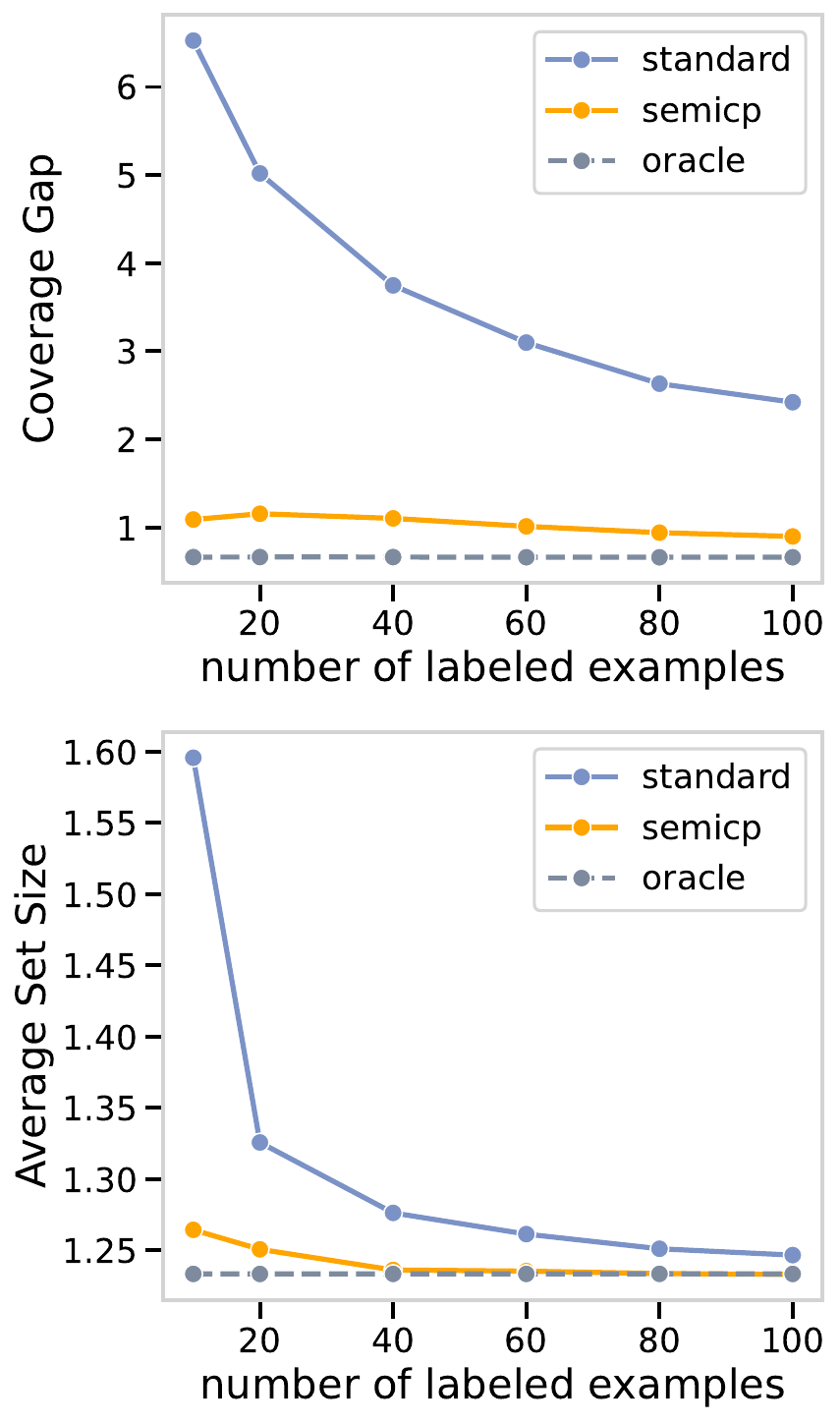}
        \caption{CIFAR-10}
    \end{subfigure}
    \begin{subfigure}{0.3\textwidth}
        \centering
        \includegraphics[width=\textwidth]{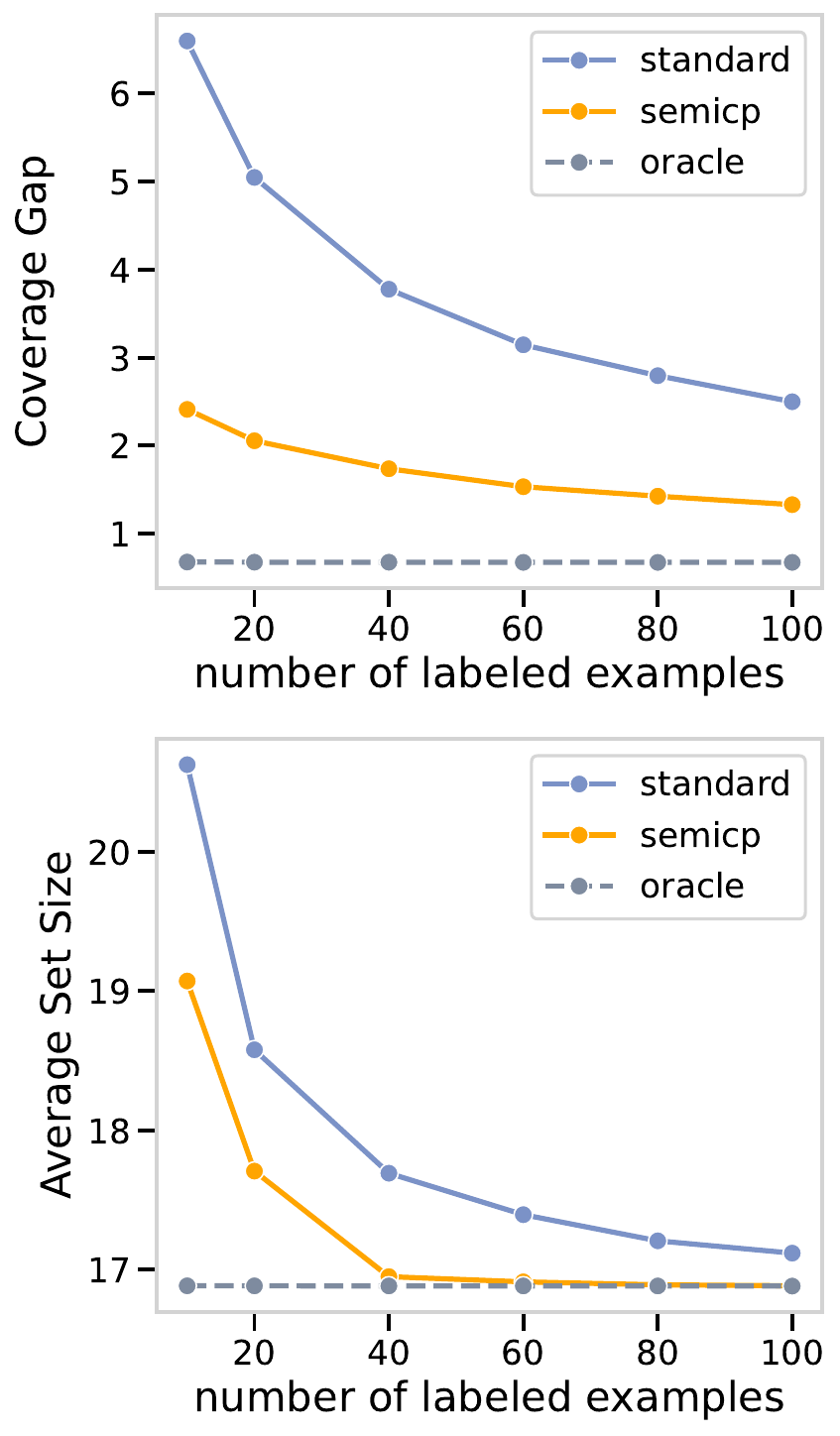}
        \caption{CIFAR-100}
    \end{subfigure}
    \begin{subfigure}{0.3\textwidth}
        \centering
        \includegraphics[width=\textwidth]{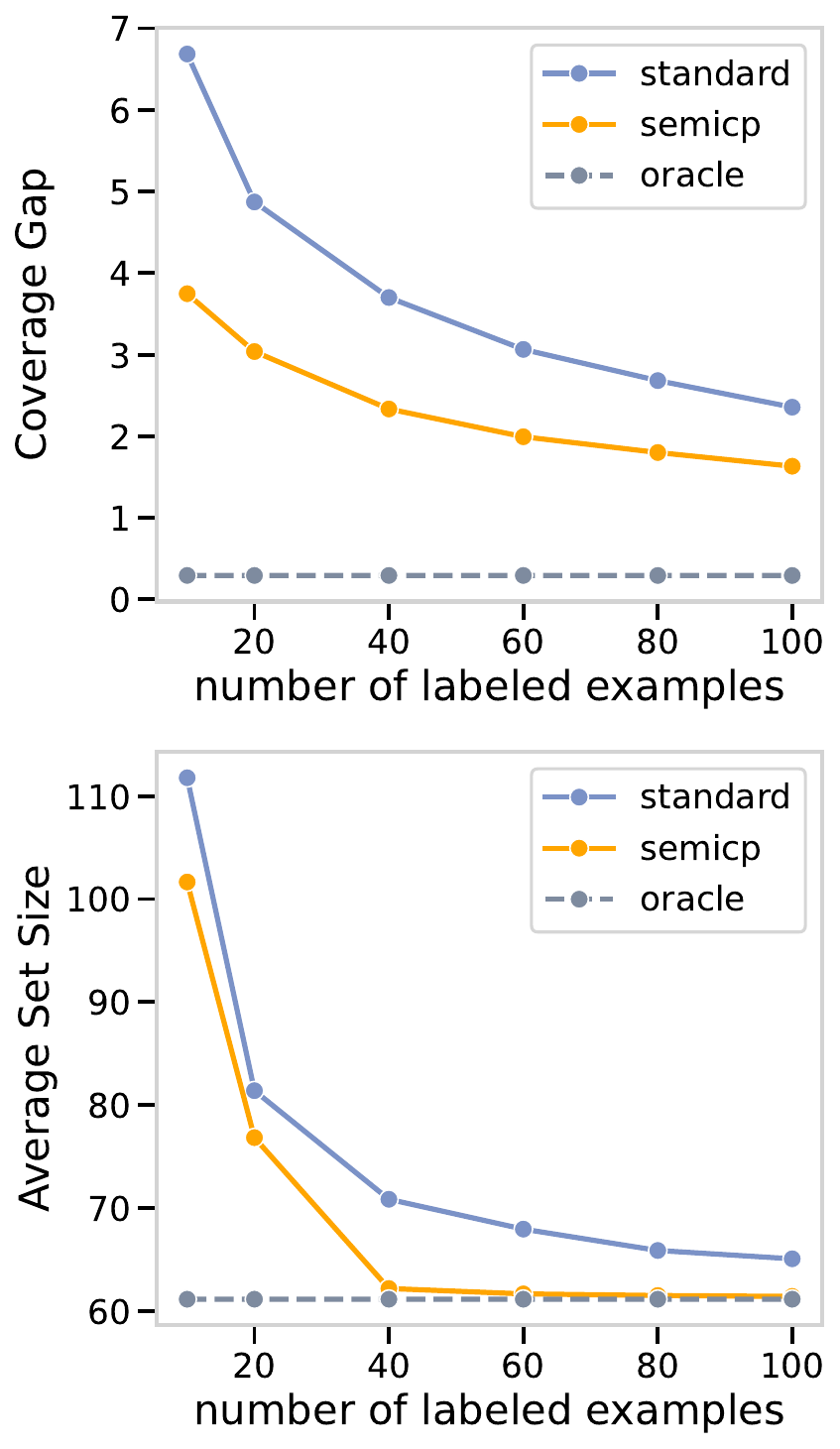}
        \caption{ImageNet}
    \end{subfigure}
    
    \vspace{-2mm}
    \caption{Average performance comparison of SemiCP with varying numbers of labeled data across three score functions on CIFAR-10, CIFAR-100, and ImageNet. 
    The results demonstrate that SemiCP can enhance the stability and efficiency of existing score functions.
    We use 4,000 unlabeled samples for CIFAR-10 and CIFAR-100, and 20,000 unlabeled samples for ImageNet.}
    \label{fig:labeled}
    \vspace{-2mm}
\end{figure*}

\begin{figure}[t]
    \centering
    \includegraphics[width=0.48\textwidth]{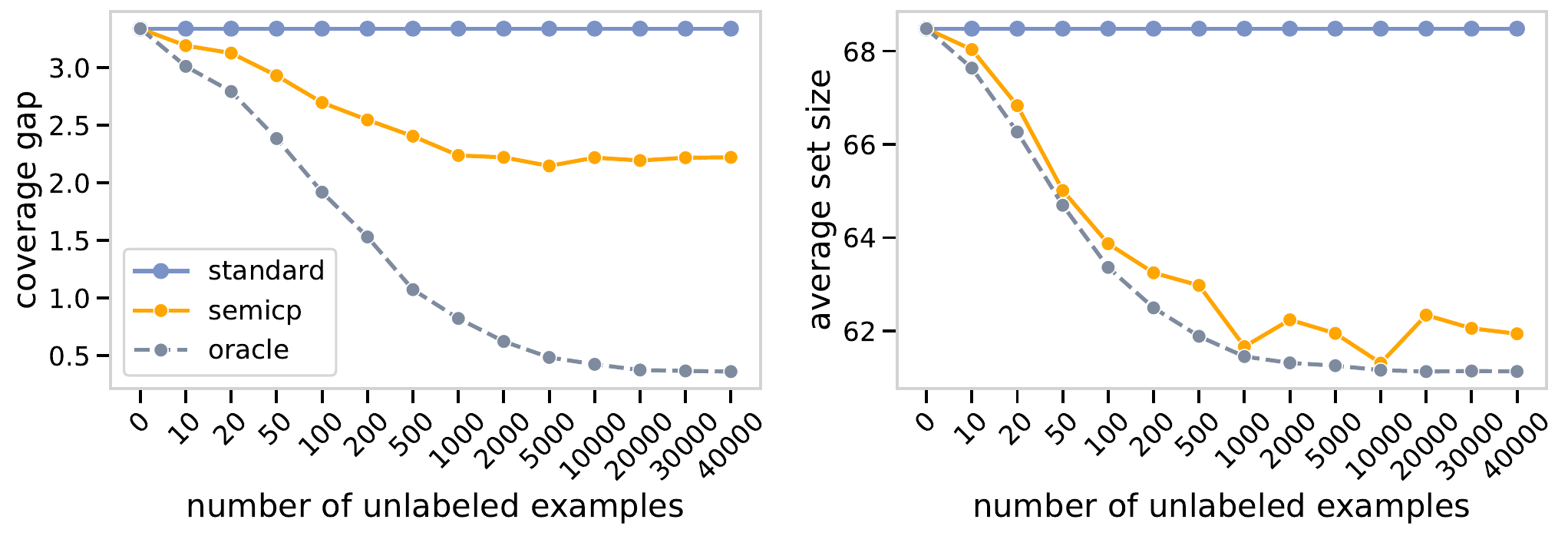} 
    \caption{Average performance comparison of SemiCP with different numbers of unlabeled data on ImageNet. 
    The number of labeled data is fixed at 50.
    } 
    \label{fig:unlabeled} 
    \vspace{-3mm}
\end{figure}

\section{Experiments}\label{sec:main_exp}
\subsection{Experimental setup}\label{sec:exp_setup}

\paragraph{Datasets} 
We evaluate our approach using three image classification datasets: CIFAR-10, CIFAR-100 \citep{krizhevsky2009learning}, and ImageNet \citep{deng2009imagenet}. Unless otherwise stated, for CIFAR-10 and CIFAR-100, we split the total of 10,000 samples into 4,000 for the labeled set, 4,000 for the unlabeled set, and 2,000 for the test set. For ImageNet, we partition 50,000 samples into 20,000 for the labeled set, 20,000 for the unlabeled set, and 10,000 for the test set. We also assess robustness of SemiCP under distribution shift; see Appendix~\ref{sec:app_robust}. 
We defer the computational overhead analysis of SemiCP to Appendix~\ref{sec:computation_overhead}. 

\paragraph{Models} 
For most of our experiments, we used ResNet50 \citep{he2016deep}. In testing the generality of our approach across different models, we evaluate 10 different architectures, including ResNet \citep{he2016deep}, Mobilenet \citep{howard2019searching}, ConvNet \citep{liu2022convnet}, EfficientNet \citep{tan2021efficientnetv2}, ResNeXt \citep{xie2017aggregated}, WideResNet \citep{zagoruyko2016wide}, ViT \citep{ dosovitskiy2020image}, MNASNet \citep{tan2019mnasnet}, and RegNet \citep{radosavovic2020designing}. For ImageNet, we leverage pre-trained classifiers from TorchVision \citep{torchvision2016}, whereas for CIFAR-10 and CIFAR-100, we retrain the classifiers using the entire training set.

\paragraph{CP baselines}  
We consider three score functions: THR, APS, and RAPS. For RAPS, we set \(k_{\text{reg}} = 2\) and \(\lambda = 0.01\). All experimental results are reported as the average performance across these three score functions, and the separate results of different score functions will be presented in detail in Appendix~\ref{sec:app_score}.
Here, we use a non-random version of APS and RAPS, as the randomization of scores can impact the accuracy of the prediction scores and influence the output. We will defer the adjustment for random factors to Appendix~\ref{sec:Random} and put the results in Appendix~\ref{sec:supexp_nnm_r}.

We present three methods in total: Standard, SemiCP, and Oracle. In the marginal coverage setting, \textbf{Standard} denotes the baseline Split CP applied to the dataset as a whole, which uses $n$ labeled data points. In the group-conditional~\citep{vovk2005algorithmic} and class-conditional~\citep{shi2013applications} settings, Standard applies CP independently to each group or class. In the group-conditional setting, we perform k-means clustering on the features of the calibration set images and divide them into 20 clusters. \textbf{SemiCP} refers to our proposed method, utilizes the same $n$ labeled data points as Standard and additionally $N$ unlabeled data points. \textbf{Oracle} represents the upper bound of CP performance by assuming access to labels of unlabeled data and running Split CP on a total of $n+N$ labeled data points. All of our experiments were repeated 1,000 trails.

\paragraph{Evaluation metrics}  
The primary metrics for evaluating prediction sets are: (1) \textit{CovGap} (Average Coverage Gap), which measures the average deviation from the target coverage across runs, and (2) \textit{AvgSize} (Average Set Size), which reflects the efficiency of the prediction set. Smaller values for both metrics are desirable, though a smaller \textit{AvgSize} is meaningful only if \textit{CovGap} is small enough. This is because \textit{AvgSize} may be artificially reduced due to undercoverage, resulting in an insufficient prediction set. The "improvement" refers to the enhancement of SemiCP relative to Oracle. We further provide a detailed result of over-coverage and under-coverage metrics in Appendix~\ref{sec:supexp_covgap}. Detailed explanations of these metrics 
are provided in Appendix~\ref{sec:metric}. 

\subsection{Results}\label{sec:main_results}

\paragraph{SemiCP enhances stability and efficiency of conformal prediction.}
Fig.~\ref{fig:labeled} compares the coverage gap and average set size of our proposed SemiCP method to that of Standard across varying amounts of labeled data, and results are averaged across three score functions. The results show that SemiCP consistently achieves smaller coverage gaps and compact prediction sets. For example, on CIFAR-10 with 10 labeled samples, SemiCP reduces the average coverage gap from 6.4 to 1.1 and decreases the average set size from 1.6 to 1.27. Moreover, as the number of labeled samples increases, SemiCP’s performance steadily improves, and its efficiency gradually approaches that of the oracle. These findings indicate that our method can reliably attain $1-\alpha$ coverage while producing more efficient prediction sets.
We further compare NNM with some variants and simple alternatives, and the experimental results are reported in Appendix~\ref{sec:exp}.

\paragraph{SemiCP is consistently effective with varying amounts of unlabeled data.}
Fig.~\ref{fig:unlabeled} shows how the average coverage gap and prediction set size of SemiCP evolve as more unlabeled data are incorporated. We find that increasing the amount of unlabeled data steadily shrinks the coverage gap, bringing SemiCP’s performance closer to the Oracle and underscoring the value of unlabeled samples. At the same time, the average set size decreases markedly, demonstrating that SemiCP not only improves coverage but also yields more concise prediction sets. Notably, our method remains effective even with very few unlabeled examples. For instance, when \(n=50\) and \(N=10\), SemiCP still reduces the coverage gap by approximately 0.1 and the average set size by about 0.2, highlighting its robustness and broad applicability.

\begin{figure}[t]
    \centering
    \includegraphics[width=0.48\textwidth]{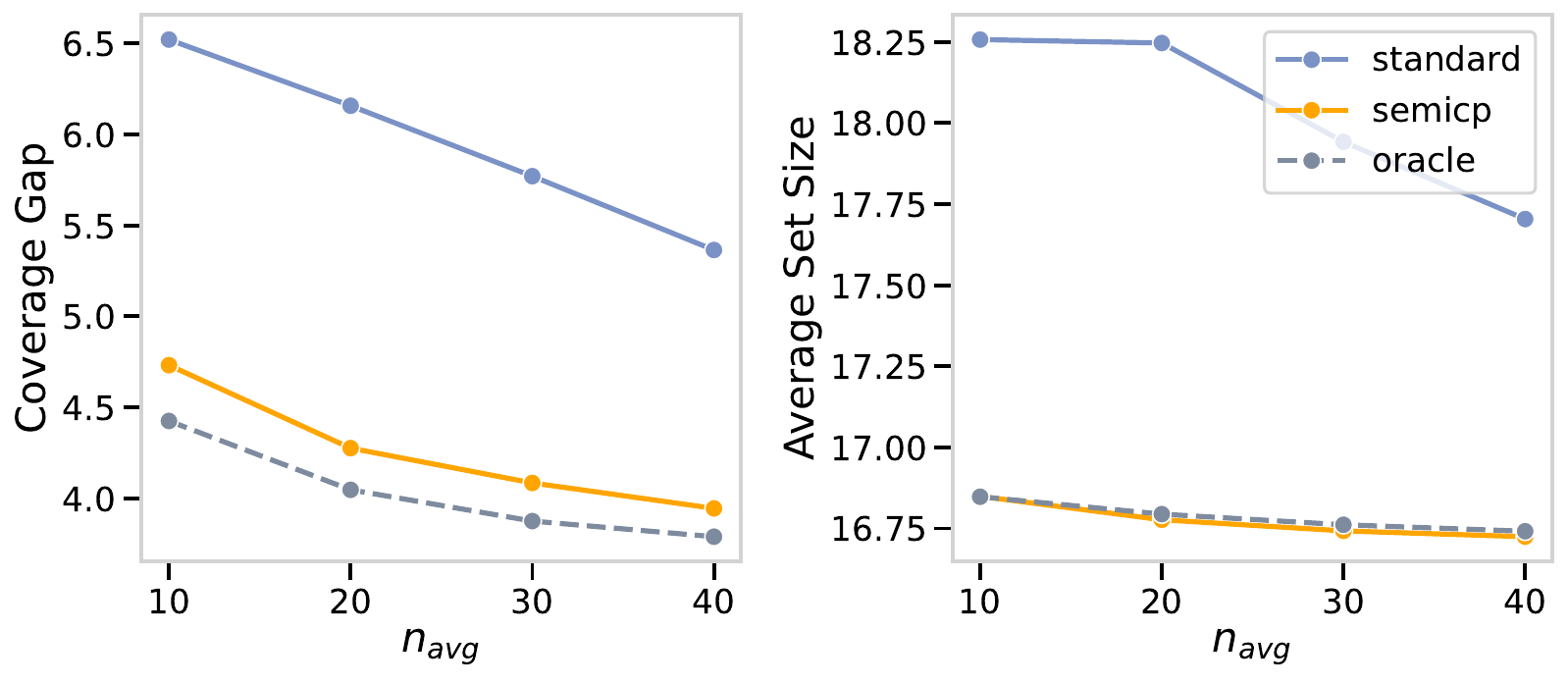}
    
    \vspace{-3mm}
    \caption{
    Average performance of SemiCP on CIFAR-100 under group conditional settings. 
    The number of unlabeled samples is set identically to that in Figure~\ref{fig:labeled}. 
    }
    
    \label{fig:conditional}
    \vspace{-3mm}
\end{figure}

\paragraph{SemiCP generalizes effectively to conditional conformal prediction.}
Fig.~\ref{fig:conditional} reports the coverage gap and average set size of SemiCP when generalized to conditional conformal prediction on CIFAR100. Here, the horizontal axis \(n_{\mathrm{avg}}\) denotes the average number of labeled examples per group or class. 
In the settings of class conditions and group conditions, SemiCP consistently outperforms the standard conformal method and even improves more than in the marginal coverage guarantee setting. For example, under class‐conditional with \(n_{\mathrm{avg}} = 10\), SemiCP reduces the coverage gap from 7.75 to 6.29 and shrinks the average set size from 18.9 down to 17, drawing remarkably close to the oracle’s performance. 
These results demonstrate that SemiCP is not only effective in marginal coverage guarantee but also provides better stability and efficiency in conditional coverage guarantee. 

\begin{figure*}[t]
    \centering
    \begin{subfigure}{0.47\textwidth}
        \centering
        \includegraphics[width=\textwidth]{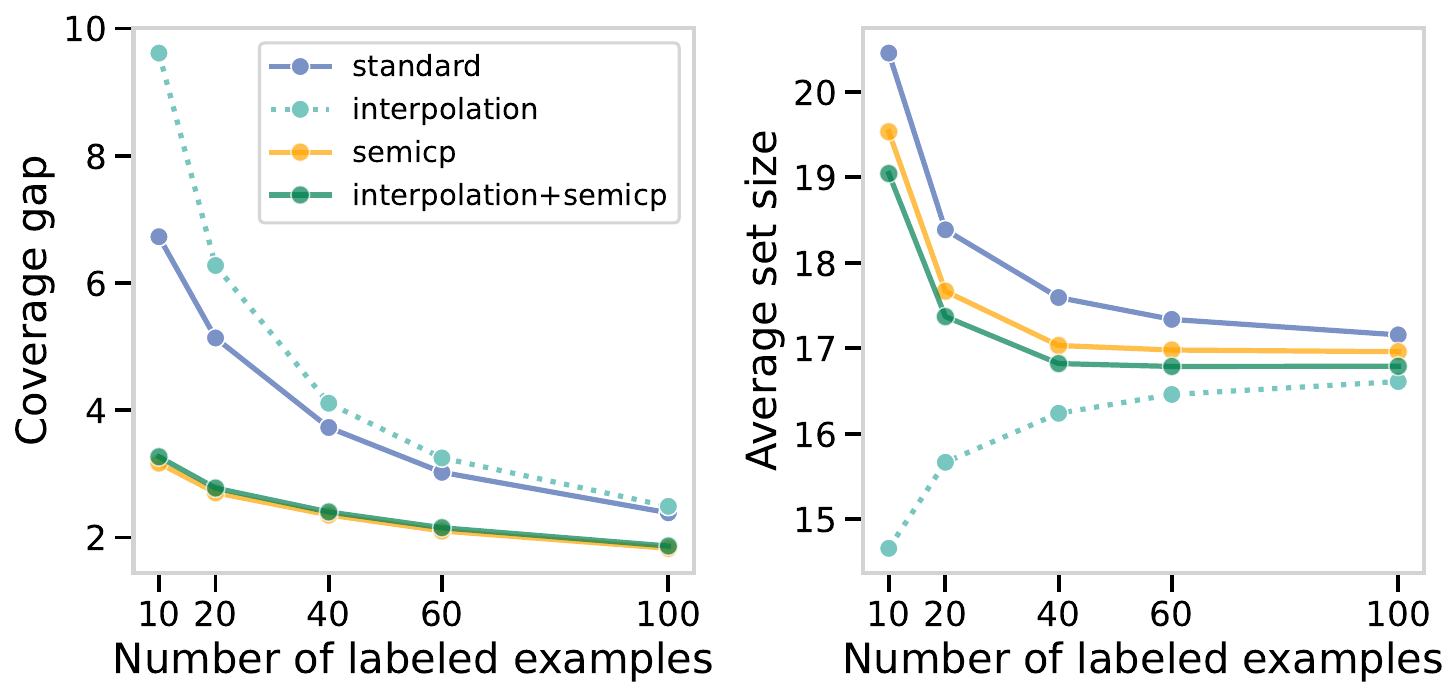}
        \caption{SemiCP+interpolation}
    \end{subfigure}
    \begin{subfigure}{0.47\textwidth}
        \centering
        \includegraphics[width=\textwidth]{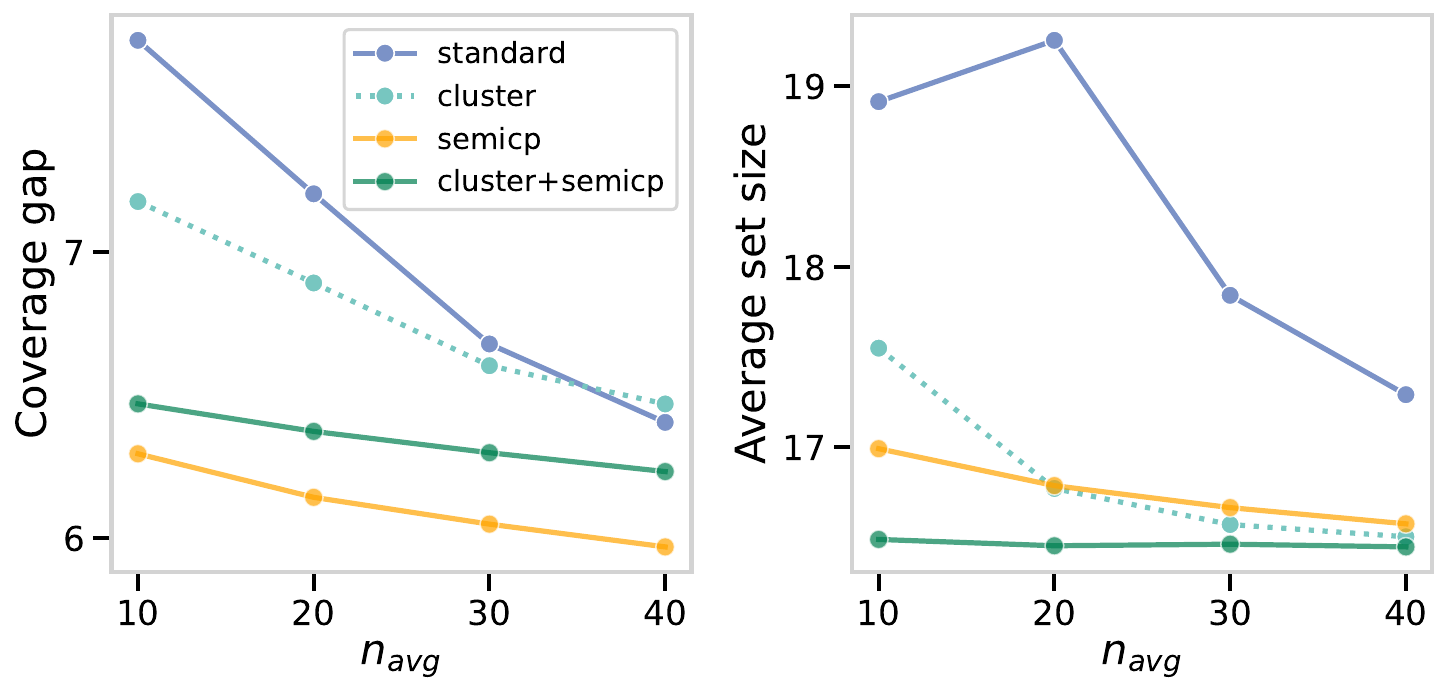}
        \caption{SemiCP+ClusterCP}
    \end{subfigure}
    
    \vspace{-1mm}
    \caption{Average performance for the integration of SemiCP and (a) Interpolation and (b) ClusterCP, conducted on CIFAR100 using ResNet50 with three score functions.
    The number of unlabeled samples is set identically to that in Figure~\ref{fig:labeled}. 
    }
    \label{fig:adaptive}
    \vspace{-1mm}
\end{figure*}

\begin{figure*}[t]
    \centering
    \includegraphics[width=0.88\linewidth]{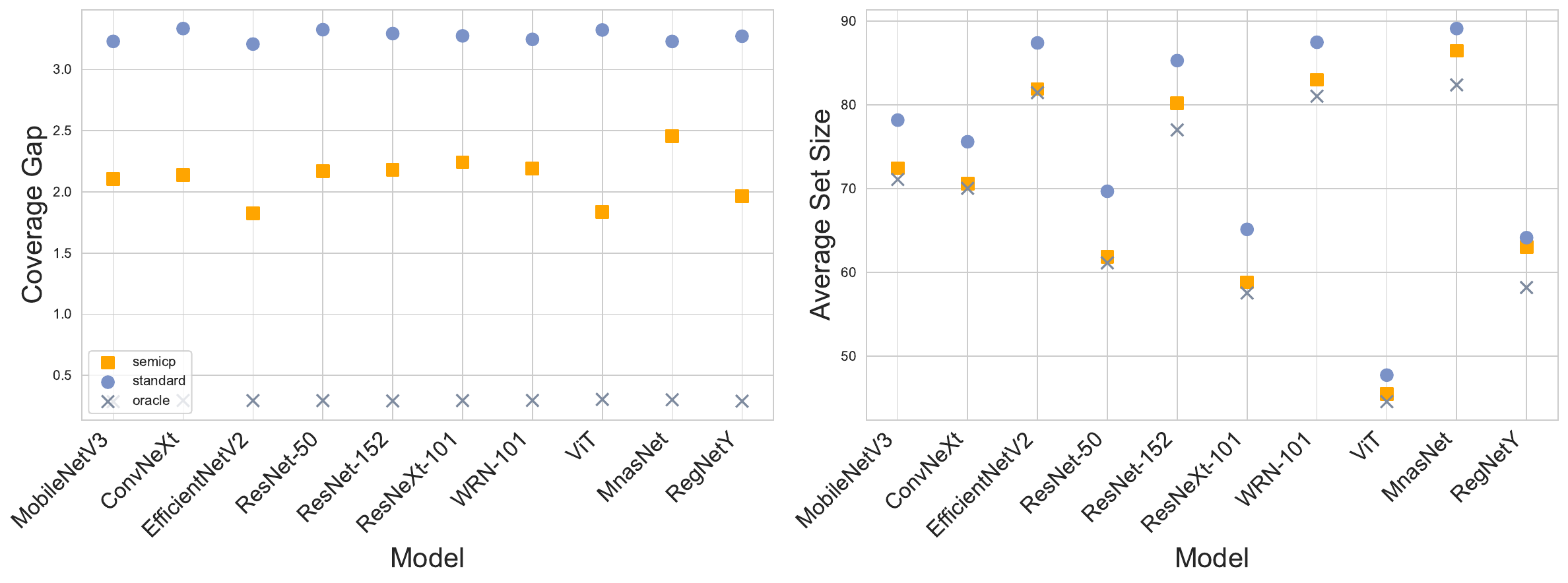}
    \vspace{-1mm}
    \caption{Average Coverage Gap and Set Size of SemiCP with different model architectures on ImageNet.}
    \label{fig:models}
    \vspace{-4mm}
\end{figure*}

\vspace{-2mm}
\paragraph{SemiCP can benefit from interpolation and ClusterCP.}
Some methods have been proposed to address its inherent trade‐offs between stability and efficiency—examples include Interpolation, which refines the coverage threshold via linear interpolation around the \((1-\alpha)\)-quantile when calibration data are scarce \citep{johansson2015handling}, and ClusterCP, which groups classes by similar score distributions to compute group‐specific thresholds for improved conditional coverage \citep{ding2023class}. Our SemiCP framework can be seamlessly integrated with these two methods, using NNM to calculate the unlabeled scores and incorporate them into the calibration set, and using these methods to calculate the threshold, further enhancing the reliability and efficiency of the prediction set.

Fig.~\ref{fig:adaptive} demonstrates the efficacy of these integrations. 
Although Interpolation alone generates a smaller prediction set, it suffers from a larger coverage gap compared to Standard, indicating greater instability and the inability to provide an effective coverage guarantee. In contrast, integrating SemiCP with Interpolation reduces the coverage gap and average set size, achieving a decrease from 9 to 3.9 in the coverage gap when \(n = 10\) and approaching the oracle's set size for \(n > 40\).
Similarly, while ClusterCP alone improves both coverage and efficiency relative to Standard under low calibration budgets, the addition of SemiCP leads to further reductions in both average set size and coverage gap across all values of \(n_{\mathrm{avg}}\). These results confirm that SemiCP can be seamlessly integrated with existing methods to exploit unlabeled data and deliver more efficient predictive sets.

\paragraph{SemiCP works well with different model architectures.} 
Fig.~\ref{fig:models} compares the performance of SemiCP on ten diverse ImageNet architectures. 
The number of labeled data and unlabeled data is fixed at 50 and 20000. Across all these models, SemiCP reduces the mean coverage gap from 3.3 to 2.1, while simultaneously tightening the average prediction set size from 75 to 70.3, consistently outperforming the Standard method. 
These results highlight the backbones robustness of SemiCP and its ability to deliver consistent improvements in both coverage and efficiency across a wide range of architectures. 
Additionally, we provide an analysis of the impact of model accuracy on SemiCP in Appendix~\ref{sec:app_pseudo}.

\section{Conclusion} \label{sec:conclusion}
In this work, we introduced a novel paradigm, SemiCP, which leverages unlabeled data to achieve more stable and efficient prediction sets. Specifically, we approximate the true score distribution of unlabeled data using the proposed unlabeled score, Nearest Neighbor Matching (NNM). Extensive experimental results demonstrate that our approach consistently achieves coverage closer to the target and produces smaller prediction sets. Our method is simple, versatile, and straightforward to implement, featuring conditional settings and various methods for enhancing conformal prediction. We hope that SemiCP will inspire future research to further explore the role of unlabeled data in conformal prediction.

\paragraph{Limitations.} 
Our theoretical results rely on the i.i.d. assumption for both labeled and unlabeled data, which is more strict than the exchangability assumption in conformal prediction. What's more, our method is limited to classification tasks and has not been extended to regression.

\newpage
{
    \section*{Acknowledgement} 

This project was supported by the Guangdong Basic and Applied Basic Research Foundation (Grant No. 2026A1515011367). We gratefully acknowledge the support of the Center for Computational Science and Engineering at the Southern University of Science and Technology.

    \small
    \bibliographystyle{ieeenat_fullname}
    \bibliography{cite}
}

\clearpage

\newpage

\setcounter{page}{1}
\onecolumn
\begin{center}
    \textbf{\Large Semi-Supervised Conformal Prediction With Unlabeled Nonconformity Score}
\end{center}
\begin{center}
    {\Large Supplementary Material}
\end{center}
\appendix

\section{Notation and Lemma}\label{sec:lemma}

We define $s$ as the random variable of true score, $\tilde{s}$ as the random variable of naive unlabeled score. $s$ takes values between $s_{\text{min}}$ and $s_{\text{max}}$. $\tilde{s}$ takes values between $\tilde{s}_{\text{min}}$ and $\tilde{s}_{\text{max}}$.
Assume $s$ and $\tilde{s}$ have no ties.

We define NNM as follows. For any data point $q$, 
assume that the labeled data \(\{(\boldsymbol{x}_i, y_i)\}_{i=1}^n\) 
are i.i.d. of $\mathcal{P}_{\mathcal{X}\mathcal{Y}}$, and independent of $q$. We choose
$$j=\text{argmin}_i|\tilde{S}(\boldsymbol{x}_i)-\tilde{S}(q)|,$$
then $S_{\text{nnm}}(q)=\tilde{S}(q)+S(\boldsymbol{x}_j, y_j)-\tilde{S}(\boldsymbol{x}_j)$. 
We refer to the random variable of $S(\boldsymbol{x}_j, y_j)$ as $s_{\text{n}}$, and the random variable of $\tilde{S}(\boldsymbol{x}_j)$ as $\tilde{s}_{\text{n}}$. 
Then we define the random variable of $S_{\text{nnm}}(q)$ as $s_{\text{nnm}}$. 

\begin{lemma}\label{lemma}
    Assume that the distribution of the nonconformity score $S$ is continuous, and the labeled data $\{(\boldsymbol{x}_i, y_i)\}^n_{i=1}$ are i.i.d. samples from the same underlying distribution. $\mathcal{C}_{1-\alpha}(\boldsymbol{x}_{\text{test}})$ is given by split conformal prediction.Then the coverage given by the calibration set $\mathcal{D}$ follows a Beta distribution:
    $$\mathbb{P}\bigl(y_{\text{test}}\in\mathcal{C}_{1-\alpha}(\boldsymbol{x}_{\text{test}})\bigr)\mid \mathcal{D}) \sim \mathrm{Beta}(l_\alpha^n, n+1 - l_\alpha^n),$$
    where $l_\alpha^n = \lceil(n+1)(1-\alpha)\rceil$.
\end{lemma}

\begin{proof}
    Let $F_s$ be the CDF of the distribution of $s$. Define $S_i=S(\boldsymbol{x}_i, y_i)$, which are i.i.d. drawn from the distribution of $s$. Let $S_{(1)}\le...\le S_{(n)}$ be the order statistics of $S_{1}, ..., S_{n}$. In the split conformal prediction, we have $y_{\text{test}}\in\mathcal{C}_{1-\alpha}(\boldsymbol{x}_{\text{test}})$ if and only if $S(\boldsymbol{x}_{\text{test}}, y_{\text{test}})\le S_{(l_\alpha^n)}$ for $l_\alpha^n=\lceil(n+1)(1-\alpha)\rceil$, so we have 
    \begin{align*}
        \mathbb{P}\bigl(y_{\text{test}}\in\mathcal{C}_{1-\alpha}(\boldsymbol{x}_{\text{test}})\bigr)\mid \mathcal{D}\bigr)&=\mathbb{P}\bigl(S(\boldsymbol{x}_{\text{test}}, y_{\text{test}})\le S_{(l_\alpha^n)}\mid \mathcal{D}\bigr)=F(S_{(l_\alpha^n)}).
    \end{align*}

    Next, let $U_i=F(S_i)$. By Probability Integral Transform, $U_i$ is uniformly distributed in $[0,1]$. Then, by definition of the Beta distribution, the $l_\alpha^n$-th order statistic from $n$ i.i.d. uniform random variables has a distribution
    $$U_{(l_\alpha^n)}\sim\text{Beta}(l_\alpha^n,n+1-l_\alpha^n).$$

    As a result, 
    $$\mathbb{P}\bigl(y_{\text{test}}\in\mathcal{C}_{1-\alpha}(\boldsymbol{x}_{\text{test}})\bigr)\mid \mathcal{D}) \sim \mathrm{Beta}(l_\alpha^n, n+1 - l_\alpha^n),$$
    where $l_\alpha^n = \lceil(n+1)(1-\alpha)\rceil$.
\end{proof}

\section{Proofs}\label{sec:proof}

\subsection{Proof of Theorem~\ref{thm:coverage}}
\begin{proof}\label{sec:proof_thm1}
Assume $s$ follows the distribution $D_{s}$, and $\tilde{s}$ follows the distribution $D_{\tilde{s}}$. Let $D_{{\text{SemiCP}}}$ be a mixture distribution of $D_s$ and $D_{\tilde{s}}$. The probability density function of $D_{\text{SemiCP}}$ is given by 
$$f_{\text{SemiCP}}=\frac{n}{n+N}f_{s}+\frac{N}{n+N}f_{{\tilde{s}}},$$
where $f_{s}$ and $f_{{\tilde{s}}}$ are the probability density function of $D_s$ and $D_{\tilde{s}}$.
We refer to an auxiliary score function as $S_{\text{SemiCP}}$, and its random variable as $s_{\text{SemiCP}}$. 
Let $B$ be an independent Bernoulli random variable
\[
B \sim \mathrm{Bernoulli}\!\left(\frac{n}{n+N}\right),
\]
and define the mixed-score random variable
\[
S_{\text{SemiCP}} =
\begin{cases}
S, & B = 1, \\
\tilde{S}, & B = 0 .
\end{cases}
\]
As a result, we have 
$$F_{\text{SemiCP}}=\frac{n}{n+N}F_{s}+\frac{N}{n+N}F_{{\tilde{s}}},$$
where $F_{\text{SemiCP}}$, $F_{s}$, $F_{{\tilde{s}}}$ are the CDFs of $s_{\text{SemiCP}}$, $s$, $\tilde{s}$.
As SemiCP uses $n$ labeled samples and $N$ unlabeled samples to calibrate, we can see that $D_{\text{SemiCP}}$ is actually the distribution of scores in SemiCP, and $\hat{\tau}_{\text{SemiCP}}$ is the $\frac{\lceil(n+N+1)(1-\alpha)\rceil}{n+N}$-quantile of the distribution $D_{\text{SemiCP}}$. 
Note that $S_{\text{SemiCP}}$ is introduced as an auxiliary quantity for the proof; it is not used in the actual method. 

Now we have
\begin{align*}
    &\quad\mathbb{P}\bigl(y_{\mathrm{test}}\in\mathcal C_{\mathrm{SemiCP}}(\boldsymbol{x}_{\mathrm{test}})\bigr)=\mathbb{P}\bigl(S(\boldsymbol{x}_{\mathrm{test}}, y_{\mathrm{test}})\le\hat{\tau}_{\text{SemiCP}}\bigr)\\
    &=\mathbb{P}\bigl(S(\boldsymbol{x}_{\mathrm{test}}, y_{\mathrm{test}})\le\hat{\tau}_{\text{SemiCP}}\bigr) - \mathbb{P}\bigl(S_{\text{SemiCP}}(\boldsymbol{x}_{\mathrm{test}})\le\hat{\tau}_{\text{SemiCP}}\bigr) + \mathbb{P}\bigl(S_{\text{SemiCP}}(\boldsymbol{x}_{\mathrm{test}})\le\hat{\tau}_{\text{SemiCP}}\bigr)\\
    &=F_s(\hat{\tau}_{\text{SemiCP}})-F_{\text{SemiCP}}(\hat{\tau}_{\text{SemiCP}}) + \mathbb{P}\bigl(S_{\text{SemiCP}}(\boldsymbol{x}_{\mathrm{test}})\le\hat{\tau}_{\text{SemiCP}}\bigr)\\
    &=F_s(\hat{\tau}_{\text{SemiCP}})-\bigl(\frac{n}{n+N}F_{s}(\hat{\tau}_{\text{SemiCP}})+\frac{N}{n+N}F_{{\tilde{s}}}(\hat{\tau}_{\text{SemiCP}})\bigr)+\mathbb{P}\bigl(S_{\text{SemiCP}}(\boldsymbol{x}_{\mathrm{test}})\le\hat{\tau}_{\text{SemiCP}}\bigr)\\
    &=\epsilon_{n,N}+\mathbb{P}\bigl(S_{\text{SemiCP}}(\boldsymbol{x}_{\mathrm{test}})\le\hat{\tau}_{\text{SemiCP}}\bigr)\\
    &\ge 1-\alpha+\epsilon_{n,N}
\end{align*}
where $\epsilon_{n,N}=\frac{N}{N+n}(F_s(\hat{\tau}_\text{SemiCP})-F_{\tilde{s}}(\hat{\tau}_\text{SemiCP}))$.
\end{proof}

\subsection{Proof of Theorem~\ref{thm:distribution}.}

\begin{proof}\label{sec:proof_thm2}
By lemma~\ref{lemma}, we have 
\begin{align*}
    \frac{\lceil(n+N+1)(1-\alpha)\rceil}{n+N+1}-\sqrt{\frac{\text{log}(\frac{2}{\delta})}{2(n+N)}}&\le\mathbb{P}\bigl(S_{\text{SemiCP}}(\boldsymbol{x}_{\mathrm{test}})\le\hat{\tau}_{\text{SemiCP}}\mid\mathcal{D}\bigr)\le\frac{\lceil(n+N+1)(1-\alpha)\rceil}{n+N+1}
    +\sqrt{\frac{\text{log}(\frac{2}{\delta})}{2(n+N)}},
\end{align*}
with probability $1-\delta$, here we set $S$ in lemma~\ref{lemma} as $S_{\text{SemiCP}}$, and $\frac{\lceil(n+N+1)(1-\alpha)\rceil}{n+N+1}$ is the expectation of the Beta distribution $\mathrm{Beta}(l_\alpha^{n+N}, n+N+1 - l_\alpha^{n+N})$. 

As illustrated in the proof of theorem~\ref{thm:coverage}, the coverage using SemiCP satisfies the following:
$$\mathbb{P}\bigl(y_{\mathrm{test}}\in\mathcal C_{\mathrm{SemiCP}}(\boldsymbol{x}_{\mathrm{test}})\mid \mathcal{D}\bigr)=\epsilon_{n,N}+\mathbb{P}\bigl(S_{\text{SemiCP}}(\boldsymbol{x}_{\mathrm{test}})\le\hat{\tau}_{\text{SemiCP}}\mid\mathcal{D}\bigr).$$

Then
\begin{align*}
    \frac{\lceil(n+N+1)(1-\alpha)\rceil}{n+N+1} &+ \epsilon_{n,N}-\sqrt{\frac{\text{log}(\frac{2}{\delta})}{2(n+N)}}\le\mathbb{P}(y_{\text{test}} \in \mathcal{C}_{\text{SemiCP}}(\boldsymbol{x}_{\text{test}})|\mathcal{D})\le\frac{\lceil(n+N+1)(1-\alpha)\rceil}{n+N+1}+\epsilon_{n,N}+\sqrt{\frac{\text{log}(\frac{2}{\delta})}{2(n+N)}},
\end{align*}
with probability $1-\delta$, that is, 
\begin{align*}
    1-\alpha&+ \epsilon_{n,N}-\sqrt{\frac{\text{log}(\frac{2}{\delta})}{2(n+N)}}\le\mathbb{P}(y_{\text{test}} \in \mathcal{C}_{\text{SemiCP}}(\boldsymbol{x}_{\text{test}})|\mathcal{D})\le1-\alpha+\frac{1}{n+N+1}+\epsilon_{n,N}+\sqrt{\frac{\text{log}(\frac{2}{\delta})}{2(n+N)}},
\end{align*}
with probability at least $1-\delta$.
\end{proof}

\subsection{Proof of Theorem~\ref{thm:nnm_asy}}\label{sec:proof_thm3}
\begin{assumption}\label{asp1}
    The probability density function $f_{\tilde{s}}(x)$ is continuous and does not vanish on the interval $[\tilde{s}_{\text{min}}, \tilde{s}_{\text{max}}]$.
    $f_{{s}}(x)$ is continuous and does not vanish on the interval $[{s}_{\text{min}}, {s}_{\text{max}}]$. 
    The probability density function $f_{s}(x|\tilde{s}=y)$ is $M$-Lipschitz with respect to $x$ and $y$. 
    $f_{s}(x|\tilde{s}=y)$ is uniformly bounded over $(x,y)$. 
\end{assumption}

\begin{remark}
    This is a mild assumption. In general, the probability that $\tilde{s}$ takes any value between $\tilde{s}_{\text{min}}$ and $\tilde{s}_{\text{max}}$ is nonzero, as shown in Fig~\ref{fig:aps_dist}. The Lipschitz property of $f_{s}(x|\tilde{s}=y)$ arises from the fact that its oscillation is not too drastic, which is also consistent with the typical behavior of $f_{s}(x|\tilde{s}=y)$ we observe in practice.
\end{remark}

\begin{assumption} \label{asp2}
    The random variables \( \tilde{s} \), \( \tilde{s}
    _{\text{n}}
    \), and \( s_{\text{n}}
    \) form a Markov chain
    \( \tilde{s} \rightarrow \tilde{s}_{\text{n}}
    \rightarrow s_{\text{n}}
    \), which implies that 
    \( \tilde{s} 
    \) is conditionally independent of \( s_{\text{n}}
    \) given \( \tilde{s}_{\text{n}} 
    \), i.e.
    \[
    f_{{s}
    _{\text{n}}}(\cdot|\tilde{s}
    _{\text{n}}=x,\tilde{s}=y)=f_{{s}
    _{\text{n}}}(\cdot|\tilde{s}
    _{\text{n}}=x) \quad \forall x,y \in \mathbb{R}.
    \]
\end{assumption}

\begin{remark}
    This assumption is also mild, because we can regard the conditional probability of $s_n$ given $\tilde{s}_n$ as the conditional probability of the true score $s$ of one point given the naive unlabeled score $\tilde{s}$ of this point, independent of the naive unlabeled score of the other points.
\end{remark}

\begin{proof} 
    By Assumption~\ref{asp1}, $f_{\tilde{s}}(x)$ is continuous and does not vanish on the interval $[\tilde{s}_{\text{min}}, \tilde{s}_{\text{max}}]$, then we define $\epsilon$ and $E$ as
    \begin{align*}
        \epsilon &:= \min_{t}\mathbb{P}(t-\delta\le\tilde{s}\le t+\delta)>0,\\
        E&:= \max_{t}\mathbb{P}(t-\delta\le\tilde{s}\le t+\delta)>0,\qquad\forall t \in[\tilde{s}_{\text{min}}, \tilde{s}_{\text{max}}].
    \end{align*}
    then we have 
    \begin{align*}
        \mathbb{P}(|\tilde{s}_{\text{n}}-\tilde{s}|\le \delta
        \mid \tilde{s}=b
        )&\ge\bigl(1-(1-\epsilon)^n\bigr),\\
        \mathbb{P}(|\tilde{s}_{\text{n}}
        -\tilde{s}|\le \delta
        \mid \tilde{s}=b
        )&\le\bigl(1-(1-E)^n\bigr),
        \qquad\forall b\in [\tilde{s}_{\text{min}}, \tilde{s}_{\text{max}}].
    \end{align*}
    As $f_{s}(x|\tilde{s}=y)$ is uniformly bounded over $(x,y)$, there exists $B$ s.t.
    $$f_{s}(x|\tilde{s}=y)\le B.$$

    As $s_{\text{nnm}} = \tilde{s} + s
    _{\text{n}} - \tilde{s}_{\text{n}}$, now we obtain the following
    \begin{align*}
        f_{s_{\text{nnm}}}(a|\tilde{s}=b) =& \int_{-\infty}^{+\infty}f_{\tilde{s}_{\text{n}}}(b+\alpha|\tilde{s}=b)f_{s
    _{\text{n}}}(a+\alpha|\tilde{s}_{\text{n}}=b+\alpha,\tilde{s}=b)d\alpha \\
        =&\int_{-\infty}^{+\infty}f_{\tilde{s}_{\text{n}}}(b+\alpha|\tilde{s}=b)f_{s
    _{\text{n}}}(a+\alpha|\tilde{s}_{\text{n}}=b+\alpha)d\alpha\\
        =&\int_{-\delta}^{\delta}f_{\tilde{s}_{\text{n}}}(b+\alpha|\tilde{s}=b)f_{s_{\text{n}}}(a+\alpha|\tilde{s}_{\text{n}}=b+\alpha)d\alpha \\
        & + (\int_{-\infty}^{-\delta}+\int_{\delta}^{+\infty})f_{\tilde{s}_{\text{n}}}(b+\alpha|\tilde{s}=b)f_{s_{\text{n}}}(a+\alpha|\tilde{s}_{\text{n}}
        =b+\alpha)d\alpha.
    \end{align*}

    The second equality is because of the Assumption~\ref{asp2}. By the Assumption~\ref{asp1}, we obtain the following inequality 
    $$-2M\delta\le f_{s}(a+\alpha|\tilde{s}
    =b+\alpha)-f_{s}(a|\tilde{s}
    =b)\le2M\delta\qquad\forall\alpha\in[-\delta,\delta],\forall a,b\in\mathbb{R}.$$

    We can see $f_{s_{\text{n}}}(a|\tilde{s}_{\text{n}}=b)$ as the conditional density probability of true score given the pesudo label score. As a result, we have $f_{s_{\text{n}}}(a|\tilde{s}_{\text{n}}=b) = f_{s}(a|\tilde{s}=b)$. 

    Then we have the following inequalities
    \begin{align*}
         f_{s_{\text{nnm}}}(a|\tilde{s}=b) - f_{s}(a|\tilde{s}=b) =&\int_{-\delta}^{\delta}f_{\tilde{s}_{\text{n}}}(b+\alpha|\tilde{s}=b)(f_{s}(a+\alpha|\tilde{s}
        =b+\alpha)-f_s(a|\tilde{s}
        =b))d\alpha \\
        & + (\int_{-\infty}^{-\delta}+\int_{\delta}^{+\infty})f_{\tilde{s}_{\text{n}}}(b+\alpha|\tilde{s}=b)(f_{s}(a+\alpha|\tilde{s}=b+\alpha) - f_s(a|\tilde{s}
        =b))d\alpha\\
         \ge &\int_{-\delta}^{\delta}f_{\tilde{s}_{\text{n}}}(b+\alpha|\tilde{s}=b)(-2M\delta)d\alpha \\
        & - B (\int_{-\infty}^{-\delta}+\int_{\delta}^{+\infty})f_{\tilde{s}_{\text{n}}}(b+\alpha|\tilde{s}=b)d\alpha\\
        \ge&(-2M\delta)\bigl(1-(1-E)^n\bigr)-B(1-\epsilon)^n,\\
        f_{s_{\text{nnm}}}(a|\tilde{s}=b) - f_s(a|\tilde{s}=b) \le &\int_{-\delta}^{\delta}f_{\tilde{s}_{\text{n}}}(b+\alpha|\tilde{s}=b)(2M\delta)d\alpha \\
        & + B (\int_{-\infty}^{-\delta}+\int_{\delta}^{+\infty})f_{\tilde{s}_{\text{n}}}(b+\alpha|\tilde{s}=b)d\alpha\\
        \le&2M\delta\bigl(1-(1-E)^n\bigr)+B(1-\epsilon)^n, \qquad \forall b \in [\tilde{s}_{\text{min}}, \tilde{s}_{\text{max}}].
    \end{align*}

    As a result, we have 
    $$\bigl(1-(1-E)^n\bigr)\bigl(-2M\delta\bigr)-B(1-\epsilon)^n\le f_{s_{\text{nnm}}}(a|\tilde{s}=b)-f_{s}(a|\tilde{s}=b)\le\bigl(1-(1-E)^n\bigr)2M\delta+B(1-\epsilon)^n.$$
    Then
    $$\bigl[\bigl(1-(1-E)^n\bigr)\bigl(-2M\delta\bigr)-B(1-\epsilon)^n\bigr]\int_{-\infty}^{+\infty}f_{\tilde{s}}(b)db\le f_{s_{\text{nnm}}}(a)-f_{s}(a)\le\bigl[\bigl(1-(1-E)^n\bigr)2M\delta+B(1-\epsilon)^n\bigr]\int_{-\infty}^{+\infty}f_{\tilde{s}}(b)db,$$
    $$\bigl(1-(1-E)^n\bigr)\bigl(-2M\delta\bigr)-B(1-\epsilon)^n\le f_{s_{\text{nnm}}}(a)-f_{s}(a)\le\bigl(1-(1-E)^n\bigr)2M\delta+B(1-\epsilon)^n,$$
    and
    $$F_{s_{\text{nnm}}}(t)-F_{s}(t)=\int_{\tilde{s}_{\text{min}}+s_{\text{min}}-\tilde{s}_{\text{max}}}^{t}\bigl(f_{s_{\text{nnm}}}(a)-f_{s}(a)\bigr)da.$$
    Let $A=2(\tilde{s}_{\text{max}}-\tilde{s}_{\text{min}})+(s_{\text{max}}-s_{\text{min}})$, then we obtain the following
    $$A\bigl[\bigl(1-(1-E)^n\bigr)\bigl(-2M\delta\bigr)-B(1-\epsilon)^n\bigr]\le F_{s_{\text{nnm}}}(t)-F_{s}(t)\le A\bigl[\bigl(1-(1-E)^n\bigr)2M\delta+B(1-\epsilon)^n\bigr].$$
\end{proof}

\section{Detailed analysis of unlabeled nonconformity score function}\label{sec:analysis}
To understand the effectiveness of the Nearest Neighbor Matching (NNM) approach for estimating the nonconformity scores of unlabeled examples, we compare it against several baseline methods of increasing complexity:

\begin{enumerate}[left=0pt]
    \item \textbf{Naive:} Uses the nonconformity score of the pseudo-label as the prediction: 
    \[
    \tilde{S}_\text{naive}(\tilde{\boldsymbol{x}}_i, S, f) = S(\tilde{\boldsymbol{x}}_i, \hat{y}_i).
    \]

    \item \textbf{Debias:} Applies a global correction by adding the average discrepancy between the true and pseudo nonconformity scores computed on labeled data:
    \begin{align*}
        \tilde{S}_\text{debias}&(\tilde{\boldsymbol{x}}_i, D_{\text{labeled}}, S, f) = S(\tilde{\boldsymbol{x}}_i, \hat{y}_i) + \frac{1}{n} \sum_{j=1}^n \left[S(\boldsymbol{x}_j, y_j) - S(\boldsymbol{x}_j, \hat{y}_j) \right].
    \end{align*}

    \item \textbf{Random Match (RM):} Adds a bias correction using a randomly selected labeled example from the dataset:
    \begin{align*}
        \tilde{S}_\text{rm}(\tilde{\boldsymbol{x}}_i;\, D_{\text{labeled}}, S, f) = S(\tilde{\boldsymbol{x}}_i, \hat{y}) &+ S(\boldsymbol{x}_j, y_j) - S(\boldsymbol{x}_j, \hat{y}), \quad j \sim \text{Uniform}\{1,\dots,n\}.
    \end{align*}

    \item \textbf{Nearest Neighbor Matching (NNM):} Selects a labeled example whose pseudo nonconformity score is closest to that of the unlabeled input, and uses it to construct a local bias correction:
    \begin{align*}
        & \tilde{S}_\text{nnm}(\tilde{\boldsymbol{x}}_i, D_{\text{labeled}}, S, f) =S(\tilde{\boldsymbol{x}}_i, \hat{y})) + S(\boldsymbol{x}_j, y_j) - S(\boldsymbol{x}_j, \hat{y}), \notag\quad \text{where } j = \mathop{\arg\min}_{j \in \{1, \dots, n\}} \left| S(\tilde{\boldsymbol{x}}_i, \hat{y}_i) - S(\boldsymbol{x}_j, \hat{y}_j) \right|.
    \end{align*}
\end{enumerate}

\paragraph{Why \textit{naive} estimation is biased.}
The naive method estimates the nonconformity score of an unlabeled input $\tilde{\boldsymbol{x}}_i$ by evaluating the score at the model’s predicted label $\hat{y}_i$, i.e.,$\tilde{S}_{\text{naive}}(\tilde{\boldsymbol{x}}_i) = S(\tilde{\boldsymbol{x}}_i, \hat{y}_i).$ By definition, the pseudo-label $\hat{y}_i$ corresponds to the most confident class under the model, and hence its nonconformity score is typically the smallest among all possible labels.

As a result, the naive method systematically underestimates the true nonconformity score, since $\tilde{y}_i$ may not coincide with the model prediction. This underestimation leads to an underestimated quantile threshold and, consequently, a conformal prediction set that is too narrow. In turn, this violates the target coverage guarantee, as the predicted set is less likely to contain the true label.

\paragraph{Why the \textit{Debias} method is inconsistent.}
The debias method attempts to correct the naive estimate by adding the global average bias observed on the labeled calibration set:
\[\tilde{S}_{\text{debias}}(\tilde{\boldsymbol{x}}_i) = \tilde{U}_i + \bar{\Delta}_n, \quad \text{where} \quad \bar{\Delta}_n = \frac{1}{n} \sum_{j=1}^n (V_j - U_j).\]

By the law of large numbers, $\bar{\Delta}_n \xrightarrow{p} \mathbb{E}[\Delta(X)]$, the expected bias across the entire input space. However, this correction ignores the local context of $\tilde{X}_i$, and generally $\Delta(\tilde{X}_i) \ne \mathbb{E}[\Delta(X)]$, especially when the bias function $\Delta(X)$ varies spatially. As a result, $\tilde{S}_{\text{debias}}(\tilde{\boldsymbol{x}}_i)$ converges to the wrong target and is therefore inconsistent for the true score $\tilde{V}_i$.

\paragraph{Why \textit{random match} is suboptimal.}
The random match (RM) method improves upon debias by using a labeled instance $j$ drawn uniformly from a local neighborhood $\mathcal{N}_n(\tilde{\boldsymbol{x}}_i, R_n)$, and correcting the pseudo-score using its observed bias:
\[\tilde{S}_{\text{rm}}(\tilde{\boldsymbol{x}}_i) = \tilde{U}_i + \Delta_j, \quad j \sim \text{Uniform}(\mathcal{N}_n).\]

This guarantees that $\|\boldsymbol{x}_j - \tilde{\boldsymbol{x}}_i\| \to 0$, helping control the error $|V_j - \tilde{V}_i|$. However, RM does not consider the pseudo-score $\tilde{U}_i$ when selecting $j$, meaning the matched point may differ significantly in model confidence. This neglect can result in poor approximation of $\Delta(\tilde{\boldsymbol{x}}_i)$, especially in regions where the model’s bias is heterogeneous, thereby limiting RM's accuracy in finite samples.

\paragraph{Why \textit{Nearest Neighbor Matching} is superior.}
Nearest Neighbor Matching (NNM) enhances RM by selecting, within the same neighborhood, the point $j^*$ whose pseudo-score $U_j$ is closest to $\tilde{U}_i$:
\[\tilde{S}_{\text{nnm}}(\tilde{\boldsymbol{x}}_i) = \tilde{U}_i + \Delta_{j^*}, \quad j^* = \arg\min_{j \in \mathcal{N}_n} |U_j - \tilde{U}_i|.\]

This matching leverages both feature-space proximity and pseudo-score similarity, yielding a local and adaptive bias correction. Under mild continuity assumptions, we can show that $\boldsymbol{x}_{j^*} \xrightarrow{p} \tilde{\boldsymbol{x}}_i$, and thus $\Delta_{j^*} \xrightarrow{p} \Delta(\tilde{\boldsymbol{x}}_i)$, making NNM an asymptotically consistent estimator. Moreover, NNM achieves lower estimation error in finite samples by selecting a neighbor that better matches both the location and the model output, making it theoretically sound and practically more effective.

Finally, we present the experimental results of coverage gap and prediction set size for these four methods in Appendix~\ref{sec:app_bias}, demonstrating that our approach achieves the best improvement in the stability and efficiency of CP. Additionally, we provide ablation studies on the selection criteria and number of nearest neighbors in Appendix~\ref{sec:app_neigh_num} and Appendix~\ref{sec:supexp_neighbor}, respectively.

\section{SemiCP vs. Semi-supervised Quantile Estimation} \label{sec:compare}

Our approach, SemiCP, addresses the inaccuracy of quantile estimation in conformal prediction (CP) by incorporating nonconformity scores computed for each unlabeled observation into the calibration process. In contrast, semi-supervised quantile estimation methods \citep{zhang1997estimating, chakrabortty2022semi, angelopoulos2023prediction} leverage auxiliary unlabeled data directly to estimate quantiles. Below, we summarize the key innovations and advantages of SemiCP:

\begin{enumerate}
    \item \textbf{Training-free and information-efficient.} Semi-supervised quantile estimation typically requires auxiliary information \citep{zhang1997estimating}, the training of additional models \citep{chakrabortty2022semi}, or iterative optimization such as gradient descent \citep{angelopoulos2023prediction}. In contrast, SemiCP requires no extra data beyond the standard CP inputs and is entirely training-free.
    
    \item \textbf{Nonconformity-based per-point scores.} Traditional semi-supervised quantile estimation yields a single, global quantile estimate, whereas SemiCP computes a nonconformity score for each unlabeled instance, directly incorporating the core CP concept of nonconformity. This per-point scoring confers several benefits: it naturally extends to conditional conformal prediction, facilitates integration with other CP enhancements, and yields more fine-grained uncertainty measures. We demonstrate these advantages and report significant empirical gains in Section~\ref{sec:main_results}.
    
    \item \textbf{Compatibility with randomized score functions.} Many CP score functions—such as APS \citep{romano2020classification}, RAPS \citep{angelopoulos2020uncertainty}, and SAPS \citep{huang2023conformal}—incorporate randomization that cannot be captured by parametric models. Semi-supervised quantile estimation fails to adapt to these randomized scores, whereas SemiCP can easily accommodate them. In Appendix~\ref{sec:Random}, we outline the adaptation strategy, and in Appendix~\ref{sec:supexp_nnm_r} we present corresponding experimental results.
\end{enumerate}

\section{Unlabeled nonconformity score with random factor}\label{sec:Random}
In the main experiments, we employ the non-random version of the nonconformity score, as the introduction of a random factor can influence the nearest-neighbor matching process. However, some nonconformity scores, such as APS, utilize randomization techniques to get tighter prediction sets. In this section, we describe how to incorporate a random factor into the Nearest-Neighbor Matching (NNM) method.

Let the score with a random factor be denoted as \( S(\boldsymbol{x}_i, y_i, u_i) \), where \( u_i \) is an independent random variable following a uniform distribution on \([0,1]\). The randomized version of NNM requires only a slight modification to the original method: the nearest neighbor is still matched using the no-random score, but the same random factor is applied during the computation process. Specifically, it can be expressed as:
\[
\tilde{S}_\text{nnm-r}(\tilde{\boldsymbol{x}}_i, \mathcal{D}_\text{labeled}, S, f, u_i) = S(\tilde{\boldsymbol{x}}_i, \hat{y}, u_i) + S(\boldsymbol{x}_{j}, y_{j}, u_i) - S(\boldsymbol{x}_{j}, \hat{y}, u_i),
\]
\[
where \ j = \mathop{\arg\min}_{j \in \{1, ..., n\}} \left| S(\tilde{\boldsymbol{x}}_i, \hat{y}) - S(\boldsymbol{x}_{j}, \hat{y}) \right|.
\]

Here, \( u_i \) remains independent of \( (\boldsymbol{x}_i, y_i) \), but the same \( u_i \) in \(S(\cdot)\) is used for each \( \tilde{\boldsymbol{x}}_i \). The experimental results for NNM-R are presented in Appendix~\ref{sec:supexp_nnm_r}.

\section{Supplemental algorithms}\label{sec:sup_al}
The algorithm of SemiCP is Algorithm~\ref{alg:semicp}. The algorithm of SemiCP-conditional is Algorithm~\ref{alg:semicp_condition}.
\begin{algorithm}
\caption{SemiCP: Semi-supervised Conformal Prediction}
\begin{algorithmic}[1]
\Require Labeled data $\mathcal{D}_{\text{labeled}} = \{(\boldsymbol{x}_i, y_i)\}_{i=1}^n$, Unlabeled data $\mathcal{D}_{\text{unlabeled}} = \{\tilde{\boldsymbol{x}}_i\}_{i=1}^N$, Significance level $\alpha \in (0, 1)$, Score function $S(\cdot, \cdot)$, Prediction model $f(\cdot)$, Test input $\boldsymbol{x}_{\text{test}}$, Label set $\mathcal{Y}$
\Ensure Prediction set $\mathcal{C}_\text{SemiCP}(\boldsymbol{x}_{\text{test}})$

\State Compute nonconformity scores for labeled data: $s_i = S(\boldsymbol{x}_i, y_i), \quad i=1,\dots,n$
\State Estimate nonconformity scores $\tilde{s}_i$ for unlabeled data $\tilde{\boldsymbol{x}}_i$:
\For{$i = 1$ \textbf{to} $N$}
    \State $j = \mathop{\arg\min}_{j \in \{1,\dots,n\}} \left| S(\tilde{\boldsymbol{x}}_i, \hat{y})) - S(\boldsymbol{x}_j, \hat{y}) \right|$
    \State $\tilde{s}_i = S(\tilde{\boldsymbol{x}}_i, \hat{y}) + S(\boldsymbol{x}_j, y_j) - S(\boldsymbol{x}_j, \hat{y})$
\EndFor
\State Compute threshold: $\hat{\tau}_\text{SemiCP} = \text{Quantile}\left( \{s_i\}_{i=1}^n \cup \{\tilde{s}_i\}_{i=1}^N, \frac{\lceil (n+N+1)(1-\alpha) \rceil}{n+N} \right)$
\State Form prediction set: $\mathcal{C}_\text{SemiCP}(\boldsymbol{x}_\text{test}) = \{y \in \mathcal{Y} : S(\boldsymbol{x}_\text{test}, y) \leq \hat{\tau}_\text{SemiCP} \}$
\State \Return $\mathcal{C}_\text{SemiCP}(\boldsymbol{x}_{\text{test}})$
\end{algorithmic}
\label{alg:semicp}
\end{algorithm}

\begin{algorithm}
\caption{SemiCP-conditional: Semi-supervised Conformal Prediction on conditional setting}
\begin{algorithmic}[1]
\Require Labeled data $\mathcal{D}_{\text{labeled}} = \{(\boldsymbol{x}_i, y_i)\}_{i=1}^n$, Unlabeled data $\mathcal{D}_{\text{unlabeled}} = \{\tilde{\boldsymbol{x}}_i\}_{i=1}^N$, Significance level $\alpha \in (0, 1)$, Score function $S(\cdot, \cdot)$, Prediction model $f(\cdot)$, Test input $\boldsymbol{x}_{\text{test}}$, Label set $\mathcal{Y}$, Group set $\mathcal{G}$
\Ensure Prediction set $\mathcal{C}_\text{SemiCP}(\boldsymbol{x}_{\text{test}})$

\State Compute nonconformity scores for labeled data: $s_i = S(\boldsymbol{x}_i, y_i), \quad i=1,\dots,n$
\State Estimate nonconformity scores $\tilde{s}_i$ for unlabeled data $\tilde{\boldsymbol{x}}_i$:
\For{$i \in \{1,...,N\}$}
    \State $j = \arg\min_{j \in \{1,\dots,n\}} \left| S(\tilde{\boldsymbol{x}}_i, \hat{y}) - S(\boldsymbol{x}_j, \hat{y}) \right|$
    \State $\tilde{s}_i = S(\tilde{\boldsymbol{x}}_i, \hat{y}) + S(\boldsymbol{x}_j, y_j) - S(\boldsymbol{x}_j, \hat{y})$
\EndFor
\For{$G_i\in \mathcal{G}$}
\State Select all nonconformity scores belonging to group $G_i$: \(\{s_i\}_{i=1}^{n_g}\), \(\{\tilde{s}_i\}_{i=1}^{N_g}\)
\State Compute threshold: $\hat{\tau}_\text{SemiCP}^g = \text{Quantile}\left( \{s_i\}_{i=1}^{n_g} \cup \{\tilde{s}_i\}_{i=1}^{N_g}, \frac{\lceil (n_g+N_g+1)(1-\alpha) \rceil}{n_g+N_g} \right)$
\EndFor
\State Form prediction set: $\mathcal{C}_\text{SemiCP}(\boldsymbol{x}_\text{test}\in G_i) = \{y \in \mathcal{Y} : S(\boldsymbol{x}_\text{test}, y) \leq \hat{\tau}_\text{SemiCP}^g \}$
\State \Return $\mathcal{C}_\text{SemiCP}(\boldsymbol{x}_{\text{test}})$
\end{algorithmic}
\label{alg:semicp_condition}
\end{algorithm}

\section{Computational overhead of algorithm}\label{sec:computation_overhead}
\paragraph{Theoretical Complexity.}
Let $n$ denote the number of labeled calibration samples and $N$ the number of unlabeled samples. We assume that computing a nonconformity score takes constant time $\mathcal{O}(1)$.

For the \textbf{standard} split conformal predictor, calibration only uses the $n$ labeled samples, leading to a time complexity of $\mathcal{O}(n)$. The \textbf{oracle} method performs calibration on all $n+N$ samples and thus requires $\mathcal{O}(n+N)$ time.

For the proposed \textbf{SemiCP} method, the NNM step introduces additional computation. The pseudo scores of labeled samples are first sorted in $\mathcal{O}(n\log n)$ time. Then, for each of the $N$ unlabeled samples, the nearest pseudo score is found via binary search in $\mathcal{O}(\log n)$ time, resulting in $\mathcal{O}(N\log n)$ complexity for the matching step. Therefore, the overall complexity of SemiCP is
\(
\mathcal{O}(n + n\log n + N\log n + N) = \mathcal{O}((n+N)\log n).
\)
In our settings where $N \gg n$, the additional overhead of SemiCP compared to the oracle method is only a logarithmic factor, which is practically negligible.

\paragraph{Empirical Runtime.}
We empirically evaluate the calibration runtime of the three methods using the TorchCP framework~\cite{huang2025torchcp}, excluding model inference and test-time prediction. Experiments are conducted with a ResNet-50 model on ImageNet using the THR score, on a single RTX 4090 GPU, averaged over 1000 runs. As shown in Table~\ref{tab:runtime_comparison}, SemiCP incurs slightly higher runtime due to the additional nearest-neighbor matching step, but the absolute overhead remains extremely small.

\begin{table}[h]
\centering
\caption{Average calibration runtime (seconds per run). }
\label{tab:runtime_comparison}
\begin{tabular}{lccc}
\toprule
Method & Standard & Oracle & SemiCP \\
\midrule
Runtime (s) & 0.000216 & 0.000273 & 0.001583 \\
\bottomrule
\end{tabular}
\end{table}

\section{Evaluation metrics}\label{sec:metric}
\paragraph{AvgSize (Average Prediction Set Size)} 
The average prediction set size quantifies the mean number of labels in the prediction sets:
\[
\text{AvgSize} = \frac{1}{|D_{\text{test}}|} \sum_{(\boldsymbol{x}, y) \in D_{\text{test}}} |\mathcal{C}(\boldsymbol{x})|.
\]

\paragraph{CovGap (Average Coverage Gap)} 
Coverage gap measures how far the marginal coverage deviates from the desired coverage level of \(1-\alpha \). The marginal coverage is defined as 
$\hat{c} = \frac{1}{|D_{\text{test}}|} \sum_{(\boldsymbol{x}, y) \in D_{\text{test}}} \mathbb{I}_{y \in \mathcal{C}(\boldsymbol{x})},$ then the average coverage gap is given by:
\[
\text{CovGap} = 100 \times |\hat{c} - (1 - \alpha)|.
\]
For conditional conformal prediction, coverage gap evaluates the deviation of each subgroup’s coverage from the target coverage \(1 - \alpha\). For example, in class-conditional conformal prediction, let \(\mathcal{J}^y = \{i \in [N']: Y'_i = y\}\) denote the index set of samples with label \(y\). The coverage for class \(y\) is defined as:
$\hat{c}_y = \frac{1}{|\mathcal{J}^y|} \sum_{i \in \mathcal{J}^y} \mathbb{I}_{y'_i \in \mathcal{C}(\boldsymbol{x}'_i)},$
then the class-conditional coverage gap is given by:
\[
\text{CovGap} = 100 \times \frac{1}{|\mathcal{Y}|} \sum_{y \in \mathcal{Y}} |\hat{c}_y - (1 - \alpha)|.
\]

\paragraph{Improvement}
Throughout the paper, "improvement" refers to the relative performance gain of a method over a baseline. Suppose \(M\) denotes the evaluation metric of interest, the improvement is computed as:
\[
\text{improvement} = \frac{M_{\text{Standard}} - M_{\text{SemiCP}}}{M_{\text{Standard}} - M_{\text{Oracle}}} \times 100\%.
\]

\paragraph{OverCovGap and UnderCovGap} 
In section~\ref{sec:supexp_covgap}, we also report Over-Coverage Gap and Under-Coverage Gap metrics to avoid the potential masking effect of the overall coverage gap, which may conceal the disadvantage of under-coverage. The Over-Coverage Gap and Under-Coverage Gap are defined as:
\[
\text{OverCovGap} = 100 \times \mathbb{I}\{\hat{c} > (1 - \alpha)\} \cdot |\hat{c} - (1 - \alpha)|,
\]
\[
\text{UnderCovGap} = 100 \times \mathbb{I}\{\hat{c} < (1 - \alpha)\} \cdot |\hat{c} - (1 - \alpha)|.
\]

\section{Supplemental experiments}\label{sec:exp}
\subsection{Supplemental metrics with OverCovGap and UnderCovGap metrics} \label{sec:supexp_covgap}
Since the CovGap metric may obscure the separate effects of over-coverage and under-coverage, we additionally measure the over-coverage and under-coverage of our method, SemiCP, denoted as OverCovGap and UnderCovGap (see Appendix~\ref{sec:metric}). Fig.~\ref{fig:Over_under_gap} illustrates the extent to which SemiCP improves over-coverage and under-coverage compared to the standard method under varying amounts of labeled data. The results demonstrate that our method consistently achieves lower levels of both over-coverage and under-coverage, suggesting a robust improvement over the standard method.

\begin{figure*}[htbp]
    \centering
    \begin{subfigure}{0.32\textwidth}
        \centering
        \includegraphics[width=\textwidth]{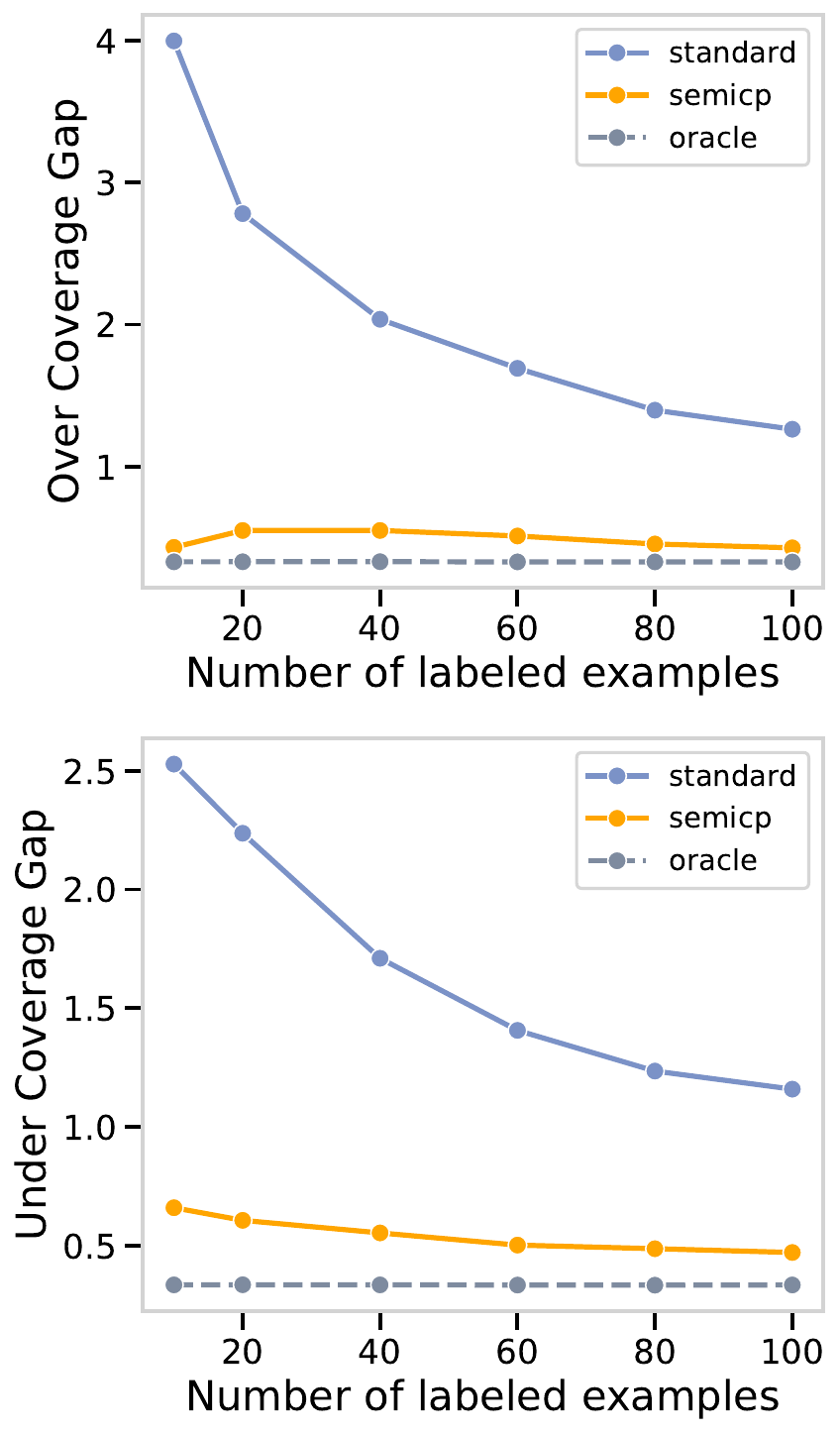}
        \caption{CIFAR-10}
    \end{subfigure}
    \begin{subfigure}{0.32\textwidth}
        \centering
        \includegraphics[width=\textwidth]{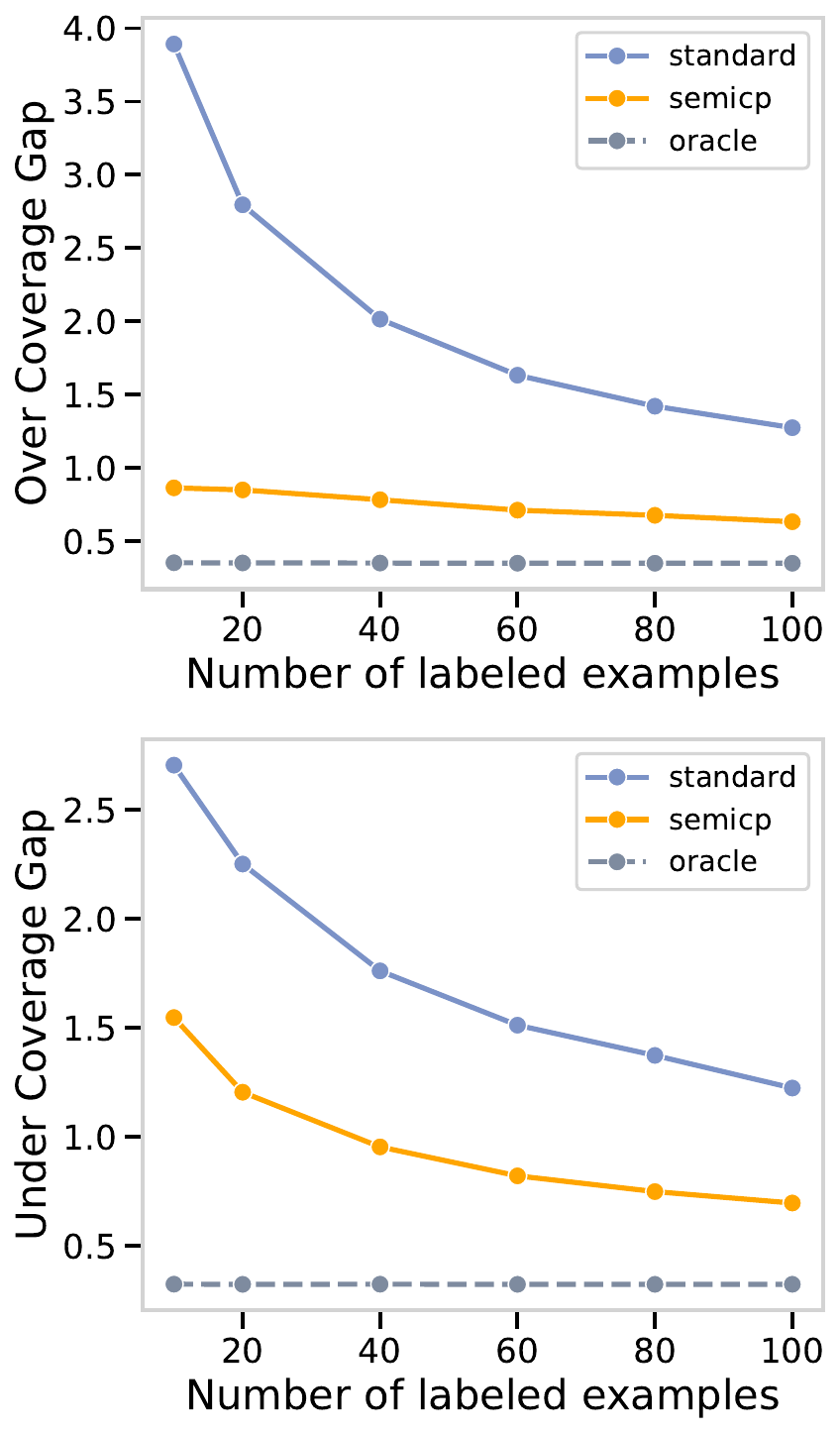}
        \caption{CIFAR-100}
    \end{subfigure}
    \begin{subfigure}{0.32\textwidth}
        \centering
        \includegraphics[width=\textwidth]{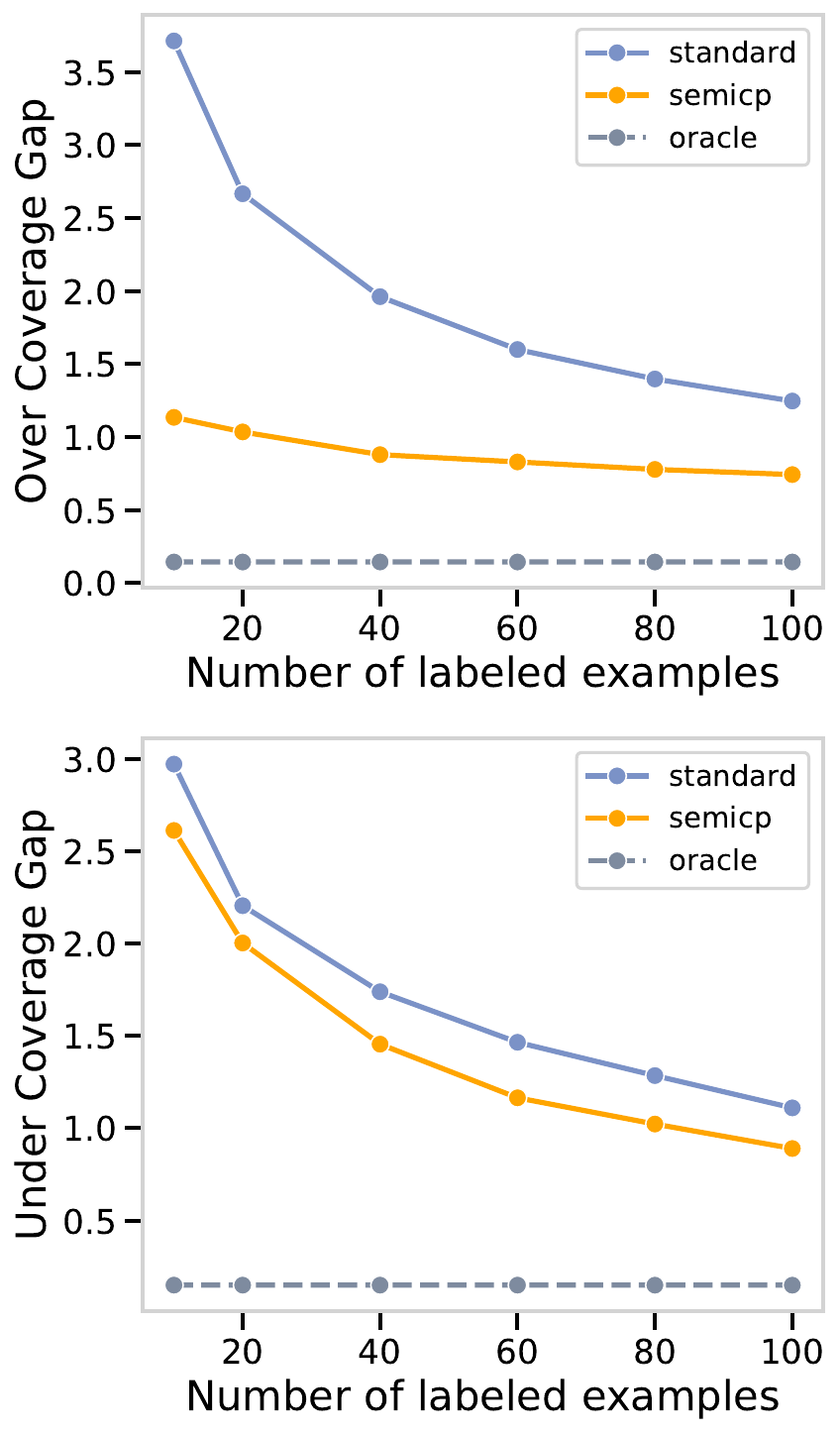}
        \caption{ImageNet}
    \end{subfigure}
    
    \caption{Comparison of the relationship between Over Coverage Gap (top) and Under Coverage Gap (bottom) with different label data numbers. Experiments conducted on CIFAR-10, CIFAR-100, and ImageNet datasets using ResNet50, average on three score functions with 1000 different trials.}
    \label{fig:Over_under_gap}
\end{figure*}

\subsection{SemiCP is compatible with various score functions} \label{sec:app_score}
Table \ref{tab:scores} summarizes the performance of our approach under three different score functions—THR, APS, RAPS—across CIFAR10, CIFAR100, and ImageNet. The results reveal that SemiCP consistently narrows the coverage gap and reduces the average set size, regardless of the labeled score employed. For instance, SemiCP reduces 81.89\% coverage gap and 97.32\% set sizes for APS, while its performance nearly matches that of the oracle method. Similar trends are observed on THR, RAPS, where SemiCP shows substantial improvements. Notably, SemiCP is more effective for those scores that initially yield poorer performance. Specifically, the improvement in APS is greater than that in RAPS, which in turn exceeds the improvement observed with THR. These improvements confirm the compatibility of SemiCP across different labeled score functions. 

\begin{table*}[h]
    \centering
    \caption{Average coverage gap and set size with different score functions on three datasets. Standard errors are reported in parentheses. The number of labeled data is fixed at 80.}
    \renewcommand\arraystretch{1.2}
    \resizebox{\textwidth}{!}{
    \setlength{\tabcolsep}{3mm}{
    \begin{tabular}{llcccccc}
        \toprule
        \multicolumn{2}{c}{\textbf{Score}} & \multicolumn{2}{c}{\textbf{THR}} & \multicolumn{2}{c}{\textbf{APS}} & \multicolumn{2}{c}{\textbf{RAPS}} \\
        \cmidrule(lr){3-4} \cmidrule(lr){5-6} \cmidrule(lr){7-8}
        
        \textbf{Dataset} & \textbf{Method} & \textbf{CovGap} & \textbf{AvgSize} & \textbf{CovGap} & \textbf{AvgSize} & \textbf{CovGap} & \textbf{AvgSize} \\
        \midrule
        \multirow{3}{*}{CIFAR10} & standard & 2.64 (2.0) & 0.91 (0.0) & 2.62 (1.9) & 1.44 (0.1) & 2.63 (1.9) & 1.40 (0.1) \\
                                  & semicp   & 0.88 (0.8) & 0.90 (0.0) & 0.96 (0.8) & 1.42 (0.0) & 0.98 (0.8) & 1.38 (0.0) \\
        \arrayrulecolor{gray}\cmidrule(lr){2-8}
                                  & oracle   & 0.65 (0.5) & 0.90 (0.0) & 0.66 (0.5) & 1.42 (0.0) & 0.66 (0.5) & 1.38 (0.0) \\
        \midrule
        \multirow{3}{*}{CIFAR100} & standard & 2.68 (2.0) & 1.21 (0.2) & 2.88 (2.1) & 39.40 (5.2) & 2.83 (2.1) & 11.02 (0.9) \\
                                  & semicp   & 2.57 (1.8) & 1.19 (0.3) & 0.75 (0.6) & 38.56 (1.2) & 0.95 (0.8) & 10.92 (0.3) \\
        \arrayrulecolor{gray}\cmidrule(lr){2-8}
                                  & oracle   & 0.69 (0.5) & 1.15 (0.0) & 0.67 (0.5) & 38.57 (0.9) & 0.67 (0.5) & 10.93 (0.2) \\
        \midrule
        \multirow{3}{*}{ImageNet} & standard & 2.69 (2.0) & 1.70 (0.5) & 2.66 (2.0) & 179.35 (51.2) & 2.70 (2.0) & 16.57 (3.4) \\
                                  & semicp   & 2.63 (1.9) & 1.62 (0.5) & 1.13 (0.8) & 166.98 (22.3) & 1.64 (1.2) & 15.90 (2.3) \\
        \arrayrulecolor{gray}\cmidrule(lr){2-8}
                                  & oracle   & 0.29 (0.2) & 1.52 (0.0) & 0.30 (0.2) & 166.15 (2.7) & 0.29 (0.2) & 15.74 (0.2) \\
        \midrule
        \multicolumn{2}{c}{\textbf{improvement}} & \textbf{32.26\%} & \textbf{55.43\%} & \textbf{81.89\%} & \textbf{97.32\%} & \textbf{71.51\%} & \textbf{96.11\%} \\
        \bottomrule
    \end{tabular}}}
    \label{tab:scores}
\end{table*}

\begin{figure*}[htbp]
    \centering
    \begin{subfigure}{0.32\textwidth}
        \centering
        \includegraphics[width=\textwidth]{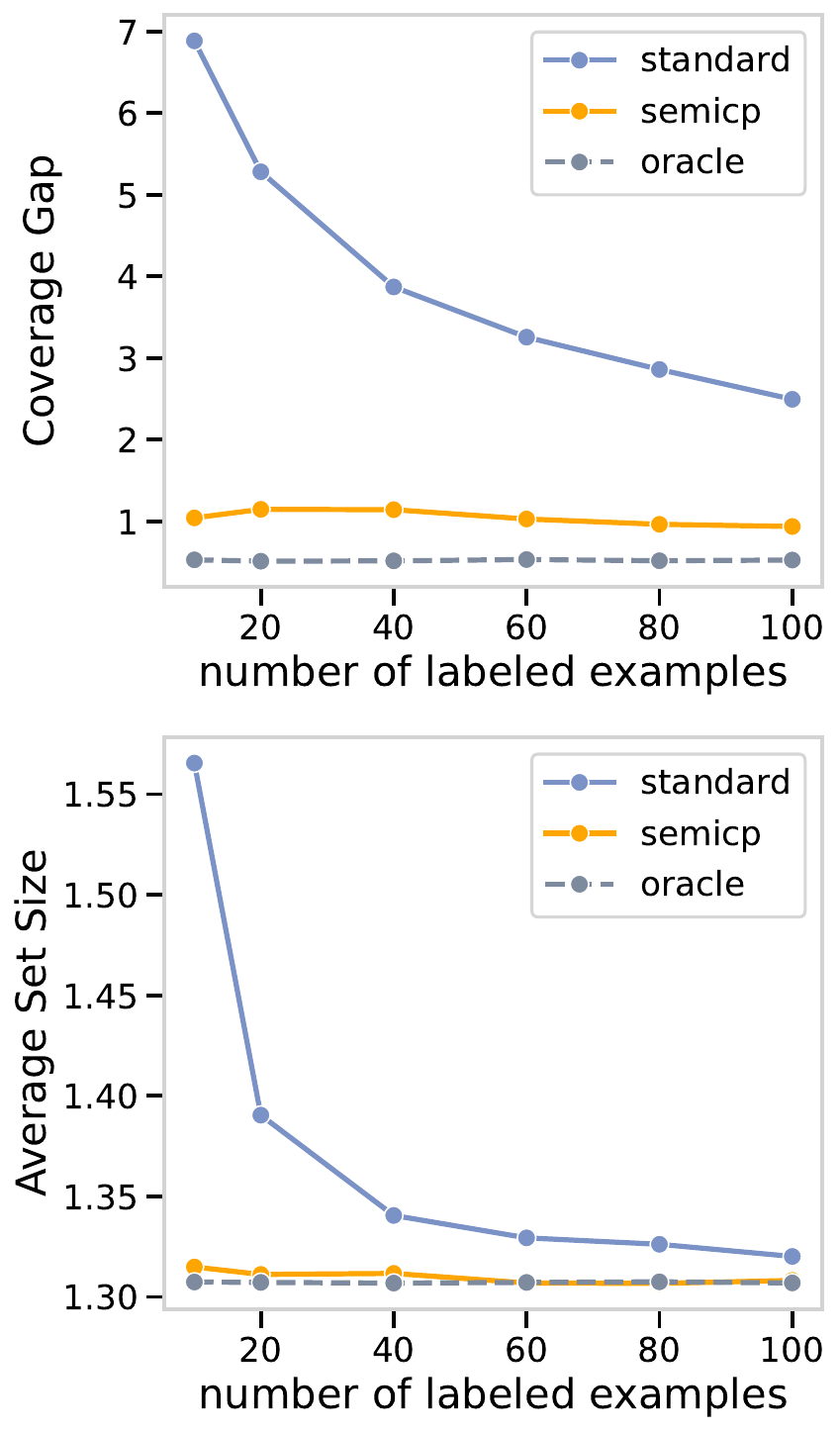}
        \caption{CIFAR-10}
    \end{subfigure}
    \begin{subfigure}{0.32\textwidth}
        \centering
        \includegraphics[width=\textwidth]{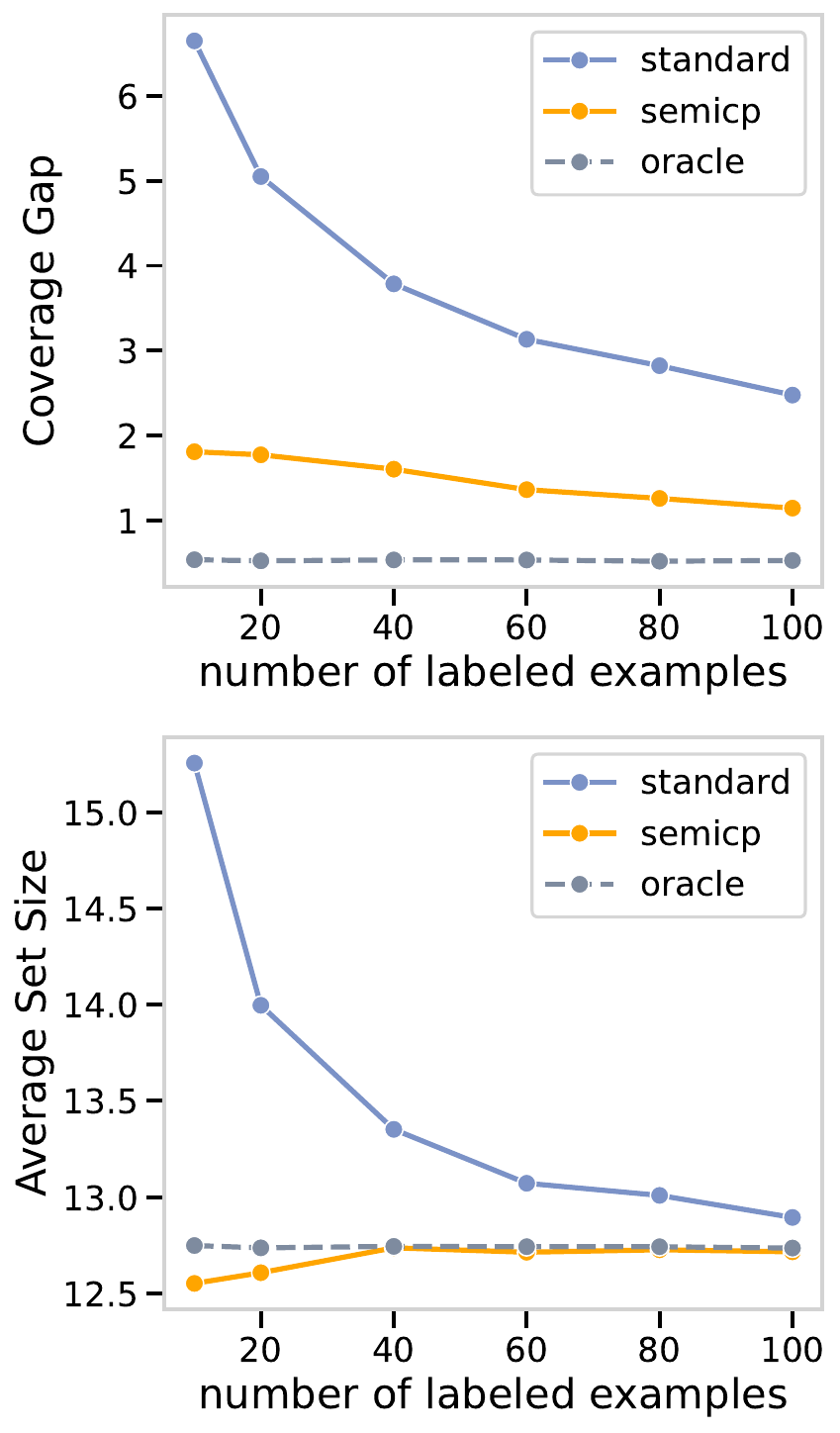}
        \caption{CIFAR-100}
    \end{subfigure}
    \begin{subfigure}{0.32\textwidth}
        \centering
        \includegraphics[width=\textwidth]{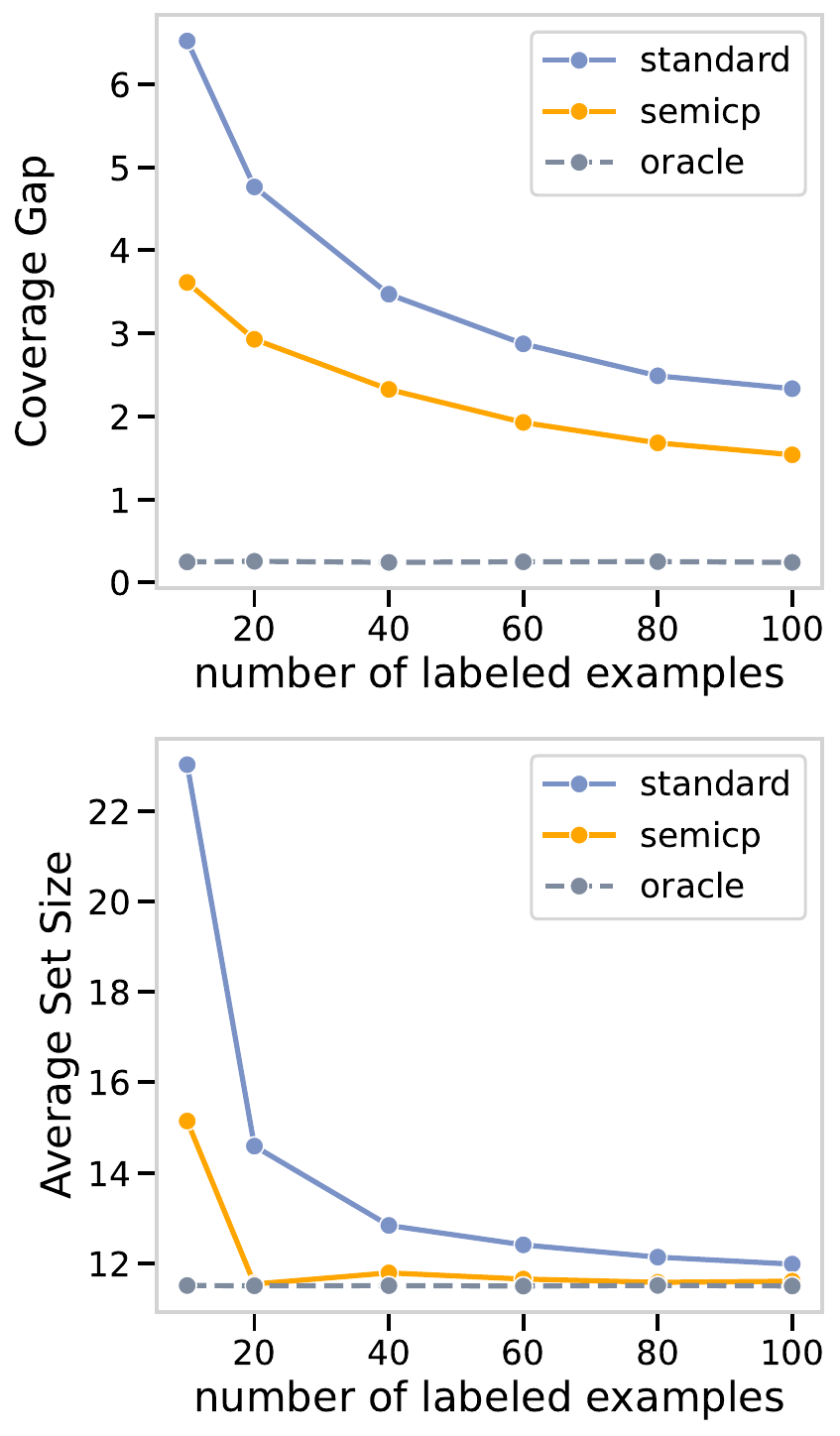}
        \caption{ImageNet}
    \end{subfigure}
    
    \caption{Average performance comparison of SemiCP with varying numbers of labeled data with SAPS on CIFAR-10, CIFAR-100, and ImageNet. We use 4,000 unlabeled samples for CIFAR-10 and CIFAR-100, and 20,000 unlabeled samples for ImageNet. The weight=0.01.}
    \label{fig:saps_result}
\end{figure*}

Fig.~\ref{fig:saps_result} shows how the coverage gap and prediction set size of SemiCP evolve as more unlabeled data are incorporated when using the SAPS score \cite{huang2023conformal}. Notably, unlike the previous experiments, we adopt the randomized version of SAPS (see~\ref{sec:Random} for details on the random factor). This is because the SAPS score function is highly non-smooth, making it difficult to obtain stable coverage guarantees without randomization. The results indicate that SemiCP effectively reduces the coverage gap and shrinks the average prediction set size with SAPS, demonstrating its strong compatibility with different score functions.

\subsection{Failure Cases of SemiCP} \label{sec:app_pseudo}
\begin{figure*}[htbp]
    \centering
    \includegraphics[width=\textwidth]{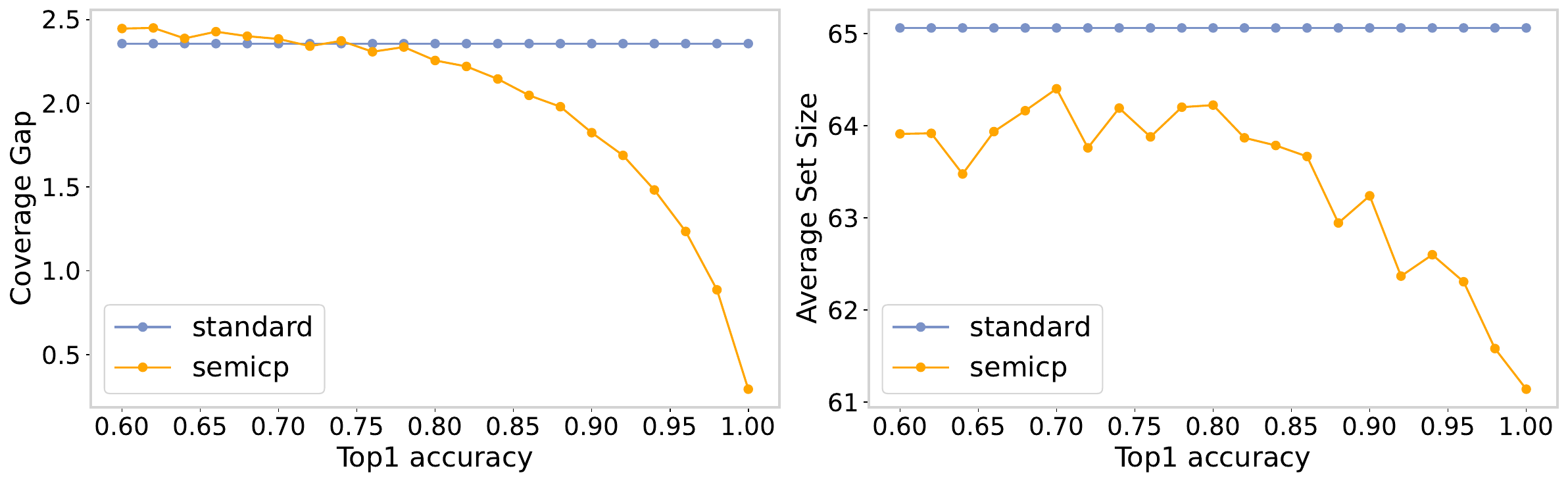} 
    \caption{Coverage gap and average set size of SemiCP with different pseudo-label accuracies. Each experiment was conducted on ImageNet and averaged over three score functions with 1000 trials. The number of labeled data and unlabeled data is fixed at 100 and 20000.} 
    \label{fig:imputeacc} 
\end{figure*}

We now turn to scenarios in which SemiCP may underperform. 
Since SemiCP selects labeled data based on pseudo-label uncertainty, its effectiveness depends critically on the quality of the pseudo-labels. 
We quantify this quality by the Top-1 accuracy of the pseudo-labeling model. 
As shown in Fig.~\ref{fig:imputeacc}, when the pseudo-label accuracy is low, SemiCP comparably to, or even worse than, the standard baseline. 
In this low-accuracy regime, unreliable pseudo-labels lead to inaccurate uncertainty estimates, which in turn distort the NNM scores and limit the performance improvement. 
As a result, the coverage gap remains large and the average set size exhibits little reduction. 
This reveals a failure case of SemiCP: when the model’s prediction accuracy is too low, the SemiCP method may fail. 


\subsection{Robustness to labeled data distribution shift}\label{sec:app_robust}
In real-world applications, the exchangeability assumption in conformal prediction often fails when labeled calibration data shift. To assess SemiCP’s robustness, we test it with labeled data from a shifted domain (ImageNet-R\citep{kornblith2019better}/ImageNet-S\citep{gao2022large}/ImageNet-V2\citep{recht2019imagenet}), while the unlabeled and test data follow the target distribution. As shown in Table~\ref{tab:shift_metrics}, SemiCP consistently outperforms standard CP in mitigating distribution shift, reducing the coverage gap by 8.35\%–49.83\% and prediction set size by 3.14\%–53.14\%, demonstrating its ability to recalibrate decision boundaries with target domain unlabeled data, even when labeled data is biased.

\begin{table*}[ht]
\centering
\caption{Comparison of CovGap and AvgSize across different distribution shift scenarios. Each experiment was conducted on ImageNet and averaged over three score functions with 100 trials. The number of labeled data and unlabeled data is fixed at 20 and 20000.}
\renewcommand\arraystretch{1.2}
    \resizebox{\textwidth}{!}{
    \setlength{\tabcolsep}{3mm}{
\begin{tabular}{ccccccc}
\toprule
& \multicolumn{2}{c}{\textbf{ImageNet-R}} 
& \multicolumn{2}{c}{\textbf{ImageNet-S}} 
& \multicolumn{2}{c}{\textbf{ImageNetV2}} \\
\cmidrule(lr){2-3} \cmidrule(lr){4-5} \cmidrule(lr){6-7}
\textbf{Method} 
& \textbf{CovGap} & \textbf{AvgSize} 
& \textbf{CovGap} & \textbf{AvgSize} 
& \textbf{CovGap} & \textbf{AvgSize} \\
\midrule
standard & 9.40 (1.1) & 546.85 (198.6) & 4.85 (3.9) & 74.73 (110.4) & 5.82 (3.9) & 91.18 (129.6) \\
semicp   & 8.64 (2.1) & 498.33 (358.9) & 3.07 (2.7) & 74.31 (132.5) & 3.06 (2.3) & 75.24 (114.9) \\
\arrayrulecolor{gray}\cmidrule(lr){1-7}
oracle   & 0.29 (0.2) & 61.19 (74.6)   & 0.29 (0.2) & 61.19 (74.7)   & 0.29 (0.2) & 61.19 (74.7) \\
\midrule
\textbf{improvement} & \textbf{8.35\%} & \textbf{9.99\%} & \textbf{38.88\%} & \textbf{3.14\%} & \textbf{49.83\%} & \textbf{53.14\%} \\
\bottomrule
\end{tabular}}}
\label{tab:shift_metrics}
\end{table*}

\subsection{Discussion of different bias estimation methods}\label{sec:app_bias}
\begin{figure*}[htbp]
    \centering
    \includegraphics[width=\textwidth]{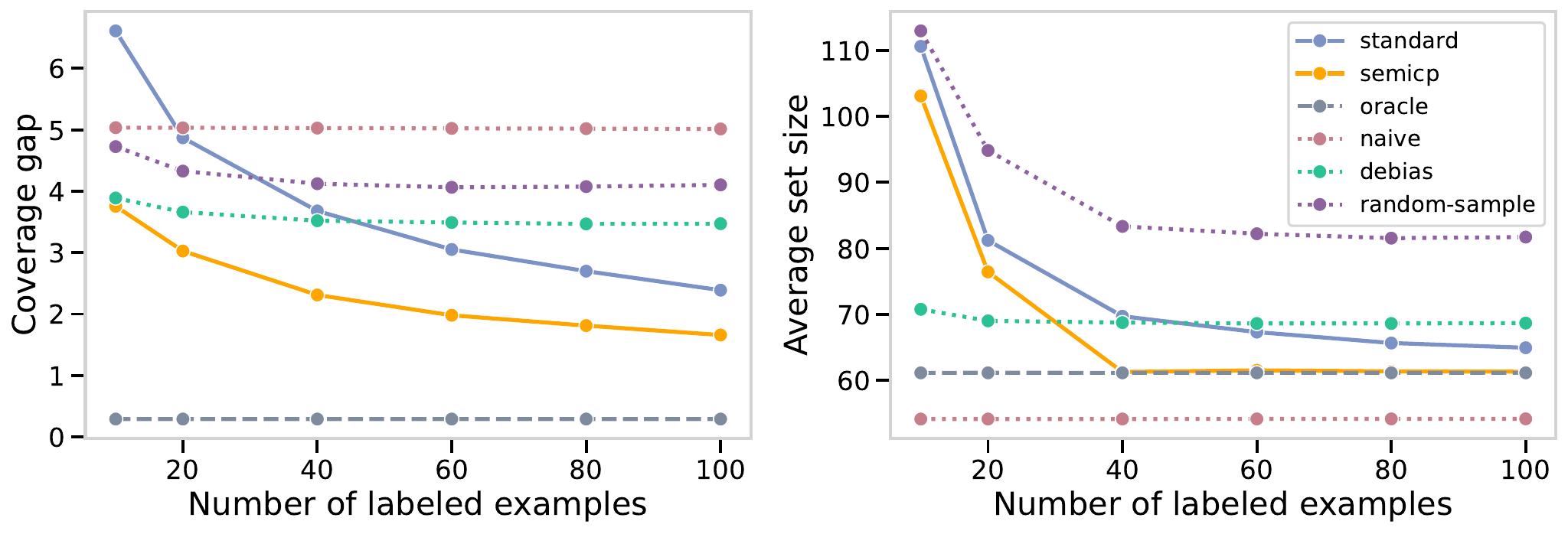} 
    \caption{Average Coverage Gap and Set Size of SemiCP with different bias estimation methods. Each experiment was conducted on ImageNet and ResNet50, averaged over three score functions with 1000 trials. The number of unlabeled data is fixed at 20000.} 
    \label{fig:debias} 
\end{figure*}

Fig.~\ref{fig:debias} compares four bias estimators. The \emph{naive} method ignores the gap between the true nonconformity score $S(\boldsymbol{x}_j,y_j)$ and the pseudo‐label score $S(\boldsymbol{x}_j,\hat{y}_j)$. The \emph{debias} estimator corrects this by using the average bias of all labeled data, $\mathrm{Bias}=\tfrac{1}{n}\sum_{j=1}^n\bigl[S(\boldsymbol{x}_j,y_j)-S(\boldsymbol{x}_j,\hat{y}_j)\bigr]$. In \emph{random‐sample}, we assign each unlabeled example the bias of a randomly chosen labeled point. Our SemiCP estimator is defined in Eq.~\ref{eq:nnm}.  

The left panel shows that SemiCP steadily reduces the coverage gap as the number of labeled samples $n$ increases, outperforming the standard CP baseline at all $n$. The other methods yield modest gains at very small $n$ (e.g.\ $n=10$) but quickly degrade relative to the baseline. The right panel plots prediction‐set size: \emph{random‐sample} always produces overly large sets, \emph{debias} performs well initially but worsens with growing $n$, and \emph{naive} underestimates scores—hence thresholds—and fails to guarantee coverage (see Section \ref{sec:UNS}). Detailed theory of four estimators is deferred to Section~\ref{sec:analysis}. Overall, SemiCP offers the best balance of reliability and efficiency.

\subsection{Ablation study of different neighborhood selection methods}\label{sec:app_neigh_num}
\begin{figure*}[t]
    \centering
    \includegraphics[width=\textwidth]{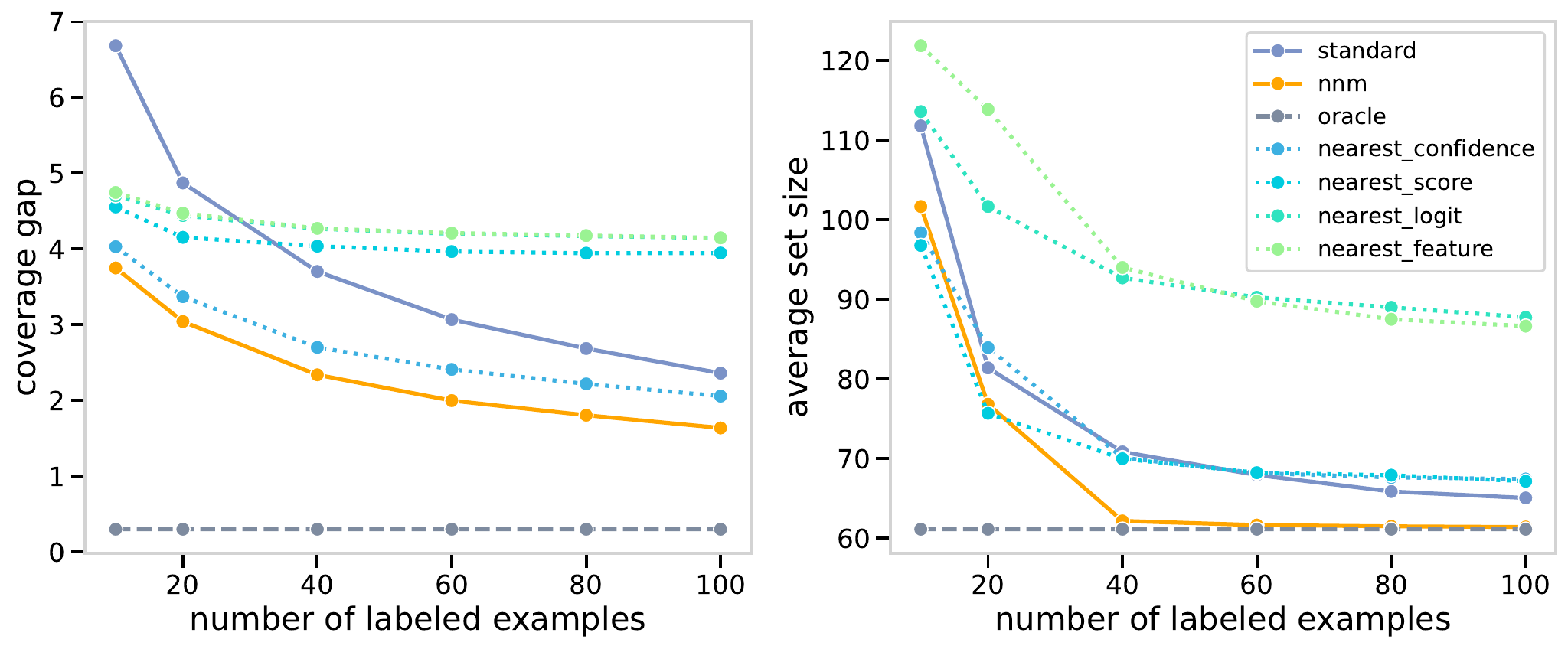} 
    \caption{Average Coverage Gap and Set Size of SemiCP with different neighborhood selection methods. Each experiment was conducted on ImageNet and ResNet50, averaged over three score functions with 1000 trials. The number of unlabeled data is fixed at 20000.} 
    \label{fig:neighbor} 
\end{figure*}

In Fig.~\ref{fig:neighbor}, we compare several neighborhood selection criteria as an ablation study. Our proposed method, \emph{nnm}, matches a labeled example with the most similar nonconformity score with a pseudo‐label. In addition, \emph{nearest\_confidence} selects labeled examples with the highest model confidence; \emph{nearest\_score} finds the neighbor whose full nonconformity‐score vector \(S(\tilde{\boldsymbol{x}})\) is most similar; \emph{nearest\_logit} matches on the raw logit outputs of the model for \(\boldsymbol{x}\); and \emph{nearest\_feature} uses the input‐feature embedding for matching. All high‐dimensional distances are measured with the Euclidean metric.  

In the coverage gap metric (left), \emph{nnm} uniformly outperforms competing methods. On average set size (right), \emph{nearest\_confidence} and \emph{nearest\_score} slightly undercut \emph{nnm} at \(n=10\), but as the number of labeled examples increases, our method \emph{nnm} exhibits a clear and sustained advantage. For example, at \(n=60\), \emph{nnm} reduces the average set size from 69 to 61, whereas all alternative methods do not improve the standard baseline. In general, these results confirm that our neighborhood‐selection criterion is optimal, providing the most reliable coverage guarantee and the smallest prediction sets.

\subsection{Ablation study of different numbers of neighborhood} \label{sec:supexp_neighbor}
\begin{figure*}[htbp]
    \centering
    \includegraphics[width=\textwidth]{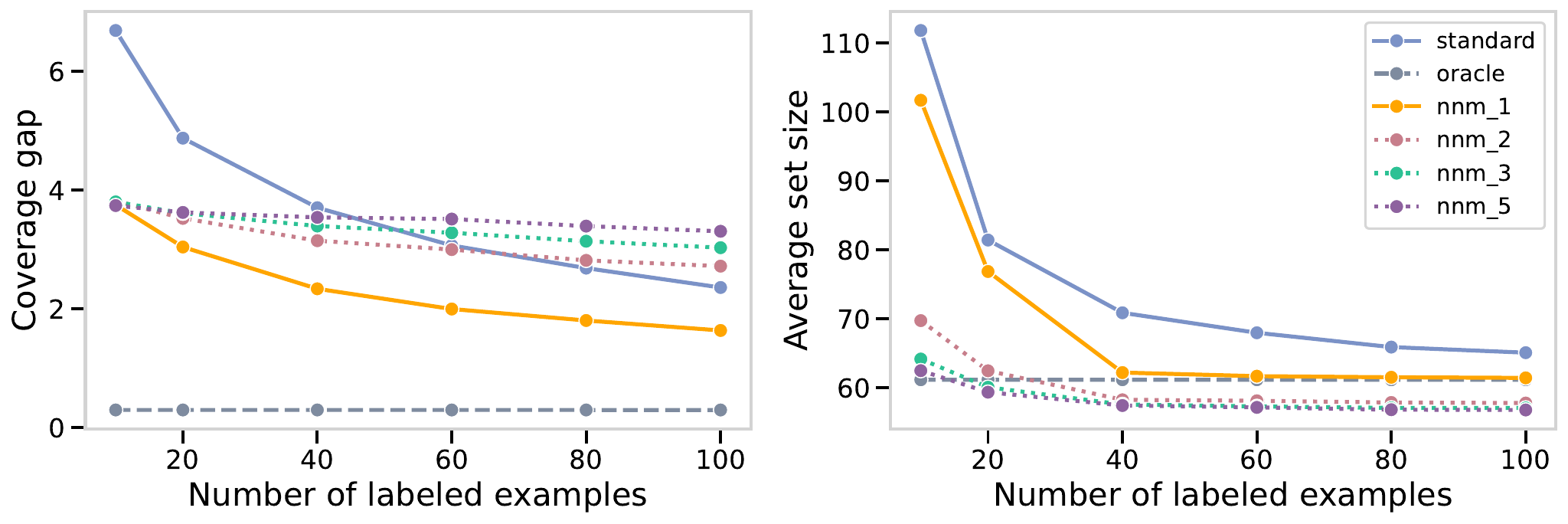} 
    \caption{Coverage gap and average set size of SemiCP with different neighborhood numbers. Each experiment was conducted on ImageNet and averaged over three score functions with 1000 trials. The number of unlabeled data is fixed at 20000.} 
    \label{fig:neighbor_num} 
\end{figure*}

In Fig.~\ref{fig:neighbor_num}, we present the results of SemiCP with different numbers of nearest neighbors. In the Fig.~\ref{fig:neighbor_num}, \emph{nnm\_k} denotes the number of nearest neighbors chosen, where $k$ indicates the number of neighbors. When multiple neighbors are selected, the bias $\Delta(\tilde{\boldsymbol{x}}_i)$ is estimated as the average of the biases of the $k$ nearest neighbors, i.e., $\frac{1}{k} \sum_{i=1}^{k} \Delta(\boldsymbol{x}_j)$. The left plot shows that with $k=1$, the coverage gap is minimized, achieving a more stable coverage of $1-\alpha$. On the right, although the set size for $k=2,3,5$ is even smaller than that of the oracle, this suggests that the coverage may not have been guaranteed, leading to overly small prediction sets. This could be because selecting closer neighbors allows for more accurate distribution estimates, resulting in better performance.

\subsection{SemiCP for nonconformity score with random factor} \label{sec:supexp_nnm_r}
Table~\ref{tab:supexp_nnm_r} compares the performance of SemiCP under nonrandom and randomized matching strategies. Both versions consistently outperform the standard method in terms of reducing coverage gap across datasets and settings, confirming the effectiveness of leveraging unlabeled data. The non-random variant achieves the lowest coverage gap, indicating that deterministic matching yields more accurate bias correction. In contrast, the randomized version produces smaller average set sizes, particularly on high-class datasets such as CIFAR100 and ImageNet. This reflects a trade-off: the random strategy sacrifices a small amount of coverage for improved compactness, which may be desirable in scenarios with large output spaces or limited labeling budgets.
\begin{table*}[h] 
    \centering
    \small
    \caption{ Average coverage gap and set size with different score functions, no random and random versions on three datasets. The number of labeled data is fixed at 100.}
    \setlength{\tabcolsep}{3pt}
    \renewcommand\arraystretch{1.1}
    \resizebox{\textwidth}{!}{
    \setlength{\tabcolsep}{3mm}{
    \begin{tabular}{c c c c c c c c c c c}
    \toprule
        \multicolumn{3}{c}{\textbf{Score}} & \multicolumn{2}{c}{\textbf{APS(no random)}} & \multicolumn{2}{c}{\textbf{APS(random)}} & \multicolumn{2}{c}{\textbf{RAPS(no random)}} & \multicolumn{2}{c}{\textbf{RAPS(random)}}\\
        \cmidrule(lr){4-5} \cmidrule(lr){6-7} \cmidrule(lr){8-9} \cmidrule(lr){10-11}
        \textbf{n} & \textbf{Dataset} & \textbf{Method} & \textbf{CovGap} & \textbf{AvgSize} & \textbf{CovGap} & \textbf{AvgSize} & \textbf{CovGap} & \textbf{AvgSize} & \textbf{CovGap} & \textbf{AvgSize} \\
    \midrule
        \multirow{9}{*}{10} & \multirow{3}{*}{CIFAR10} & standard & 6.59 & 1.94 & 6.97 & 1.34 & 6.59 & 1.83 & 6.97 & 1.33 \\
        & & semicp & 1.23 & 1.42 & 1.82 & 0.97 & 1.29 & 1.38 & 1.83 & 0.96 \\
        \arrayrulecolor{gray}\cmidrule(lr){3-11}
        & & oracle & 0.67 & 1.41 & 0.67 & 0.96 & 0.66 & 1.38 & 0.67 & 0.96 \\
        \cmidrule(lr){2-11} 
        & \multirow{3}{*}{CIFAR100} & standard & 6.04 & 46.91 & 7.82 & 15.01 & 6.13 & 12.64 & 7.67 & 6.71 \\
        & & semicp & 0.77 & 38.74 & 2.75 & 8.26 & 1.1 & 11.11 & 3.24 & 4.43 \\
        \arrayrulecolor{gray}\cmidrule(lr){3-11}
        & & oracle & 0.63 & 38.62 & 0.69 & 8.01 & 0.6 & 10.94 & 0.68 & 4.22 \\
        \cmidrule(lr){2-11} 
        & \multirow{3}{*}{ImageNet} & standard & 6.45 & 273.48 & 6.23 & 117.55 & 6.35 & 28.9 & 6.21 & 21.95 \\
        & & semicp & 1.46 & 157.77 & 4.21 & 40.13 & 3.04 & 16.2 & 4.83 & 7.8 \\
        \arrayrulecolor{gray}\cmidrule(lr){3-11}
        & & oracle & 0.3 & 166.26 & 0.31 & 36.89 & 0.26 & 15.76 & 0.32 & 6.53 \\
    \midrule
        \multirow{9}{*}{20} & \multirow{3}{*}{CIFAR10} & standard & 5.17 & 1.53 & 4.93 & 1.00 & 5.17 & 1.46 & 4.93 & 0.99 \\
        & & semicp & 1.12 & 1.41 & 1.66 & 0.96 & 1.17 & 1.38 & 1.67 & 0.95 \\
        \arrayrulecolor{gray}\cmidrule(lr){3-11}
        & & oracle & 0.67 & 1.41 & 0.66 & 0.96 & 0.65 & 1.38 & 0.66 & 0.96 \\
        \cmidrule(lr){2-11} 
        & \multirow{3}{*}{CIFAR100} & standard & 4.85 & 43.48 & 5.04 & 11.37 & 4.80 & 11.59 & 4.89 & 5.04 \\
        & & semicp & 0.70 & 38.49 & 2.34 & 7.82 & 0.98 & 10.89 & 2.62 & 4.11 \\
        \arrayrulecolor{gray}\cmidrule(lr){3-11}
        & & oracle & 0.62 & 38.63 & 0.67 & 7.98 & 0.60 & 10.94 & 0.67 & 4.21 \\
        \cmidrule(lr){2-11} 
        & \multirow{3}{*}{ImageNet} & standard & 4.25 & 219.51 & 4.88 & 67.95 & 4.46 & 20.03 & 4.65 & 11.14 \\
        & & semicp & 1.45 & 161.03 & 3.39 & 39.63 & 2.46 & 15.75 & 3.79 & 7.16 \\
        \arrayrulecolor{gray}\cmidrule(lr){3-11}
        & & oracle & 0.30 & 166.26 & 0.30 & 36.82 & 0.26 & 15.76 & 0.31 & 6.53 \\
    \midrule
        \multirow{9}{*}{50} & \multirow{3}{*}{CIFAR10} & standard & 3.62 & 1.49 & 3.42 & 0.97 & 3.60 & 1.44 & 3.42 & 0.97 \\
        & & semicp & 1.17 & 1.43 & 1.51 & 0.96 & 1.21 & 1.39 & 1.52 & 0.96 \\
        \arrayrulecolor{gray}\cmidrule(lr){3-11}
        & & oracle & 0.67 & 1.41 & 0.65 & 0.96 & 0.66 & 1.38 & 0.65 & 0.96 \\
        \cmidrule(lr){2-11} 
        & \multirow{3}{*}{CIFAR100} & standard & 3.84 & 39.96 & 3.59 & 9.57 & 3.81 & 11.07 & 3.55 & 4.67 \\
        & & semicp & 0.71 & 38.59 & 1.82 & 8.31 & 1.00 & 10.93 & 1.99 & 4.32 \\
        \arrayrulecolor{gray}\cmidrule(lr){3-11}
        & & oracle & 0.62 & 38.61 & 0.66 & 7.99 & 0.61 & 10.94 & 0.66 & 4.21 \\
        \cmidrule(lr){2-11} 
        & \multirow{3}{*}{ImageNet} & standard & 3.53 & 186.92 & 3.49 & 52.32 & 3.52 & 17.13 & 3.36 & 8.13 \\
        & & semicp & 1.35 & 165.10 & 2.42 & 43.15 & 2.03 & 15.95 & 2.73 & 7.36 \\
        \arrayrulecolor{gray}\cmidrule(lr){3-11}
        & & oracle & 0.30 & 166.25 & 0.32 & 36.68 & 0.26 & 15.76 & 0.30 & 6.51 \\
    \midrule
        \multirow{9}{*}{100} & \multirow{3}{*}{CIFAR10} & standard & 2.37 & 1.44 & 2.23 & 0.97 & 2.36 & 1.40 & 2.25 & 0.97 \\
        & & semicp & 0.99 & 1.42 & 1.23 & 0.96 & 1.00 & 1.38 & 1.23 & 0.96 \\
        \arrayrulecolor{gray}\cmidrule(lr){3-11}
        & & oracle & 0.67 & 1.41 & 0.65 & 0.96 & 0.66 & 1.38 & 0.64 & 0.96 \\
        \cmidrule(lr){2-11} 
        & \multirow{3}{*}{CIFAR100} & standard & 2.84 & 38.96 & 2.48 & 9.01 & 2.74 & 10.96 & 2.35 & 4.51 \\
        & & semicp & 0.76 & 38.64 & 1.44 & 8.20 & 0.93 & 10.93 & 1.49 & 4.24 \\
        \arrayrulecolor{gray}\cmidrule(lr){3-11}
        & & oracle & 0.63 & 38.63 & 0.66 & 8.04 & 0.61 & 10.94 & 0.68 & 4.23 \\
        \cmidrule(lr){2-11} 
        & \multirow{3}{*}{ImageNet} & standard & 2.74 & 172.76 & 2.41 & 41.71 & 2.71 & 16.24 & 2.33 & 7.00 \\
        & & semicp & 1.05 & 163.38 & 1.90 & 39.05 & 1.50 & 15.77 & 2.17 & 6.86 \\
        \arrayrulecolor{gray}\cmidrule(lr){3-11}
        & & oracle & 0.30 & 166.23 & 0.31 & 36.73 & 0.27 & 15.76 & 0.31 & 6.52 \\
    \midrule
        \multicolumn{3}{c}{\textbf{improvement}} & \textbf{85.03\%} & \textbf{100.55\%} & \textbf{55.48\%} & \textbf{88.4\%} & \textbf{74.26\%} & \textbf{101.64\%} & \textbf{48.69\%} & \textbf{88.69\%} \\
    \bottomrule
    \end{tabular}}}
    \label{tab:supexp_nnm_r}
\end{table*}

\end{document}